\documentclass[letterpaper, 10 pt, conference]{ieeeconf}  

\IEEEoverridecommandlockouts                              

\usepackage{graphicx}
\graphicspath{ {./imgs/} }
\usepackage{amsmath} 
\usepackage{amssymb}  

\usepackage{multicol}
\usepackage{multirow}
\usepackage{makecell}
\usepackage{balance}
\usepackage{setspace}

\usepackage{algorithm}
\usepackage{algpseudocode}

\usepackage{xcolor}
\usepackage{bm}

\usepackage{subfig}
\usepackage{enumerate}

\usepackage[noadjust]{cite}

\newcommand{\vect}[1]{\mathbf{\bm{#1}}}

\DeclareMathOperator*{\argmax}{arg\,max}

\title{\LARGE \bf
Efficient Extrinsic Self-Calibration of Multiple IMUs using Measurement Subset Selection
}
\author{
Jongwon Lee$^{1}$, David Hanley$^{2}$, and Timothy Bretl$^{1}$
\thanks{$^{1}$Jongwon Lee and Timothy Bretl are with the Department of Aerospace Engineering, University of Illinois Urbana-Champaign, Urbana, IL 61801, USA (Email: \texttt{\{jongwon5, tbretl\}@illinois.edu}).}
\thanks{$^{2}$David Hanley is with the School of Informatics and the Translational Healthcare Technologies Group in Centre for Inflammation Research, Institute for Regeneration and Repair, University of Edinburgh, UK (Email: {\tt\small dhanley@ed.ac.uk}).}
\thanks{This work was supported by the NASA Grant No. STTR-80NSSC20C0020.}
}

\begin{document}

\maketitle
\thispagestyle{empty}
\pagestyle{empty}

\setlength{\abovedisplayskip}{0pt} \setlength{\abovedisplayshortskip}{0pt}

\begin{abstract}


This paper addresses the problem of choosing a sparse subset of measurements for quick calibration parameter estimation. 
A standard solution to this is selecting a measurement only if its utility---the difference between posterior (with the measurement) and prior information (without the measurement)---exceeds some threshold. 
Theoretically, utility, a function of the parameter estimate, should be evaluated at the estimate obtained with all measurements selected so far, hence necessitating a recalibration with each new measurement. However, we hypothesize that utility is insensitive to changes in the parameter estimate for many systems of interest, suggesting that evaluating utility at some initial parameter guess would yield equivalent results in practice. We provide evidence supporting this hypothesis for extrinsic calibration of multiple inertial measurement units (IMUs), showing the reduction in calibration time by two orders of magnitude by forgoing recalibration for each measurement.

\end{abstract}

\section{Introduction}
\label{section:Introduction}

Inertial measurement units (IMUs) are a common component in mobile robot navigation, comprising accelerometers and gyroscopes to measure specific forces and angular rates along three orthogonal axes. While single IMUs are commonly used, deploying multiple IMUs can enhance measurement accuracy, bandwidth, and fault tolerance without increasing the overall size, weight, power, and cost~\cite{nilsson2016inertial, skog2016inertial, parsa2005estimation, wijayasinghe2018study, zhang2020lightweight, eckenhoff2020mimc}. 

To realize these benefits, accurate {\em extrinsic calibration} is crucial to determine the relative pose of each IMU to others. 
Various extrinsic calibration methods for multi-IMU systems have been developed, including those utilizing prescribed trajectories~\cite{cho2005calibration, schopp2010design, he2013novel} and aiding sensors like cameras~\cite{rehder2016extending}. 
Notably, {\em self}-calibration methods~\cite{schopp2016self,kim2017line,lee2022extrinsic}, which rely solely on IMU measurements themselves, offer significant benefits, especially in scenarios requiring recalibration due to intentional or unintentional changes in sensor configuration during robot operation, such as part loosening or thermal expansion.

Self-calibration, not bound to a predetermined trajectory, naturally experiences measurements with varying ``utility''. 
In cases like calibrating multi-IMU systems on non-rotating bodies such as spacecraft in orbit or cars on straight paths, much of the data can be low in utility.
This is especially critical for optimization-based calibration, favored for its accuracy compared to filter-based methods, as it involves processing a significant amount of low-utility (uninformative) data alongside with high-utility (informative) data during optimization. 
Therefore, identifying subsets of measurements with high utility would offer benefits such as reduced runtime and memory usage for optimization. 
\textcolor{black}{
However, this remains beneficial only if the selection process is done efficiently.
}


In this study, we present an approach to multi-IMU extrinsic calibration by selecting informative measurement subsets with high utility, significantly streamlining our previous work~\cite{lee2022extrinsic}. 
This selection is primarily based on method from existing studies~\cite{maye2013self,maye2016online} but is executed more efficiently. Specifically, we hypothesize that the measure of utility is insensitive to changes in the parameter estimate and equivalent calibration results are obtained by evaluating utility only at some initial guess of the calibration parameters. A valuable consequence of this is eliminating the need for recalibration with each new measurement, thereby reducing total computation time for calibration.

Our paper is structured as follows: 
Section~\ref{section:Related Works} reviews existing literature on informative measurement selection for self-calibration. 
Section~\ref{section:Subset Identification} describes an approach with our hypothesis for efficient measurement subset selection. 
Section~\ref{section:Methodology} delineates the solution approach for multi-IMU extrinsic self-calibration using these subsets, building on our prior work~\cite{lee2022extrinsic}.
We then proceed to assess its runtime and accuracy against benchmarks in simulation (Section~\ref{section:Simulation}) and using real-world data (Section~\ref{section:Hardware}).
Section~\ref{section:Conclusion} summarizes our findings, provides conclusions, and discuss the implications of this work.

\section{Related Works}
\label{section:Related Works}


Selecting measurements with high utility for self-calibration involves optimizing the Fisher information matrix $\vect{\mathcal{I}}$, whose inverse sets the Cramer-Rao lower bound~\cite{rao1947minimum} for parameter estimation.
Imagine a full measurement set $\mathcal{D}$ divided into $L$ segments ${\mathcal{D}^1, \dotsc, \mathcal{D}^L}$, with each segment $\mathcal{D}^l$ containing measurements across $K$ consecutive timesteps. The goal is to identify a subset of $\mathcal{D}$
\begin{equation*}
\begin{aligned}
    & \underset{\mathcal{D}^{info} \subset \mathcal{D},\, \text{card}(\mathcal{D}^{info}) = \kappa}{\argmax}
    & & f \left[ \vect{\mathcal{I}}(\mathcal{D}^{info}) \right] \\
    & \text{subject to } & & \kappa < \text{card}(\mathcal{D}),
\end{aligned} \tag{original}
\label{eq:original}
\end{equation*}   
where $f$ maps the input matrix to a scalar value (termed ``information'' in what follows), with $\text{card}(\cdot)$ denoting set cardinality and $\kappa$ a predetermined subset size. This combinatorial optimization problem, aimed at maximizing information, is typically addressed through two alternative ways in the existing literature. One approach involves choosing segments with the highest information:
\begin{enumerate}
    \item Evaluate $f[\vect{\mathcal{I}}(\cdot)]$ for segments $\mathcal{D}^1$ through $\mathcal{D}^L$.
    \item Choose $M$ segments showing the largest $f[\vect{\mathcal{I}}(\cdot)]$.
\begin{equation*}
\tag{M-largest selection}
\label{eq:M-largest}
\end{equation*}
\end{enumerate}
Schneider \textit{et al.}~\cite{schneider2019observability} and Lv \textit{et al.}~\cite{lv2022observability} used this approach in camera-IMU and LiDAR-IMU calibration, respectively.
The second approach uses a greedy algorithm, which selectively adds candidate segments $\mathcal{D}^{new}$ in the informative subset $\mathcal{D}^{info}$ if their utility exceeds threshold $\lambda$:
\begin{enumerate}
    \item Initialize $\mathcal{D}^{info}$ as an empty set.
    \item Iterate over candidate segments $\mathcal{D}^{new}$ from $\mathcal{D}^1$ to $\mathcal{D}^L$.
    \item Add $\mathcal{D}^{new}$ in $\mathcal{D}^{info}$ if $f[\vect{\mathcal{I}}(\mathcal{D}^{info}, \mathcal{D}^{new})] - f[\vect{\mathcal{I}}(\mathcal{D}^{info})] > \lambda$.
    \item Continue until all segments are processed. 
\begin{equation*}
\tag{greedy algorithm}
\label{eq:Greedy}
\end{equation*}
\end{enumerate}
Maye \textit{et al.} initially applied this approach in rangefinder-odometer~\cite{maye2013self} and later in camera-IMU calibration~\cite{maye2016online}.

Both approaches succeed in calibration only if they correctly identify informative subsets.
We show in Section~\ref{subsection:Limitations of Existing Methods for Self-Calibration with Informative Subset Selection} an edge case where \ref{eq:M-largest} fails.
\textcolor{black}{
Therefore, outside of that section, we refrain from discussing \ref{eq:M-largest}, focusing instead solely on \ref{eq:Greedy}.
}

However, the usefulness of the \ref{eq:Greedy} depends on the selection efficiency. We show that the original \ref{eq:Greedy} by Maye \textit{et al.}~\cite{maye2013self,maye2016online} may not significantly improve or might even worsen runtime for our multi-IMU calibration scenario in concern. This led us to seek improvements detailed in Section~\ref{section:Subset Identification}, where we apply our hypothesis for a more efficient execution of the greedy algorithm.

\section{Efficient Identification of Informative Measurement Subsets for Self-Calibration}
\label{section:Subset Identification}

\begin{figure*}
   \begin{minipage}[t]{0.48\textwidth}
   \begin{algorithm}[H]
   \caption{Pseudocode for calibration via the original greedy algorithm~\cite{maye2013self,maye2016online}}
   \label{alg:maye}
   \begin{algorithmic}[1]
       \footnotesize

       \State \textbf{Input: } uncalibrated parameters $\vect{\theta}^{0}$, 
       \State \textbf{Output: } calibrated parameters $\vect{\theta}^{\ast}$
       
       \State $\mathcal{D}^{info} \gets \varnothing$ 
       \State $\widehat{\vect{\theta}}^{-} \gets \vect{\theta}^{0}$
       
       \For {$new \gets 1$ to $L$} 
   
       \State {$\widehat{\vect{\theta}}^{+} \gets \texttt{Calibrate}(\mathcal{D}^{info}, \mathcal{D}^{new}; \widehat{\vect{\theta}}^{-})$}
       \State {$\vect{\mathcal{I}}(\mathcal{D}^{info},\mathcal{D}^{new}) \gets \vect{J}(\mathcal{D}^{info})|_{\widehat{\vect{\theta}}^{+}}^T \vect{J}(\mathcal{D}^{info})|_{\widehat{\vect{\theta}}^{+}} + \vect{J}(\mathcal{D}^{new})|_{\widehat{\vect{\theta}}^{+}}^T \vect{J}(\mathcal{D}^{new})|_{\widehat{\vect{\theta}}^{+}}$}
       \State {$\vect{\Sigma}_{\vect{\Theta}|\mathcal{D}^{info},\mathcal{D}^{new}} \gets \big[ \vect{\mathcal{I}}(\mathcal{D}^{info},\mathcal{D}^{new})^{-1} \big]_{\vect{\Theta} \vect{\Theta}} $}
       
       \If {$\frac{1}{2} \textrm{log} \frac{ \big| \vect{\Sigma}_{\vect{\Theta}|\mathcal{D}^{info}} \big| }{ \big| \vect{\Sigma}_{\vect{\Theta}|\mathcal{D}^{info},\mathcal{D}^{new}} \big| } > \lambda$ \textrm{or} $\mathcal{D}^{info} == \varnothing$}
       \State {$\mathcal{D}^{info} \gets \mathcal{D}^{info} \cup \mathcal{D}^{new}$}   
       \State {$ | \vect{\Sigma}_{\vect{\Theta}|\mathcal{D}^{info}} | \gets | \vect{\Sigma}_{\vect{\Theta}|\mathcal{D}^{info},\mathcal{D}^{new}} | $}
       \State {$\widehat{\vect{\theta}}^{-} \gets \widehat{\vect{\theta}}^{+} $ }
   
       \EndIf
   
       \EndFor
       \State {$\vect{\theta}^{\ast} \gets \widehat{\vect{\theta}}^{-} $}
       \State {\textbf{return} $\vect{\theta}^{\ast}$}
   \end{algorithmic}
   \end{algorithm}
   \end{minipage}\hfill
   \begin{minipage}[t]{0.48\textwidth}
   \begin{algorithm}[H]
   \caption{Pseudocode for calibration via greedy algorithm with utility evaluation at initial calibration parameters}
   \label{alg:ours}
   \begin{algorithmic}[1]
       \footnotesize

       \State \textbf{Input: } uncalibrated parameters $\vect{\theta}^{0}$, 
       \State \textbf{Output: } calibrated parameters $\vect{\theta}^{\ast}$

       \State $\mathcal{D}^{info} \gets \varnothing$ 

       \For {$new \gets 1$ to $L$} 
       
       \State {$\vect{\mathcal{I}}(\mathcal{D}^{info},\mathcal{D}^{new}) \gets \vect{J}(\mathcal{D}^{info})|_{\vect{\theta}^{0}}^T \vect{J}(\mathcal{D}^{info})|_{\vect{\theta}^{0}} + \vect{J}(\mathcal{D}^{new})|_{\vect{\theta}^{0}}^T \vect{J}(\mathcal{D}^{new})|_{\vect{\theta}^{0}}$}
       \State {$\vect{\Sigma}_{\vect{\Theta}|\mathcal{D}^{info},\mathcal{D}^{new}} \gets \big[ \vect{\mathcal{I}}(\mathcal{D}^{info},\mathcal{D}^{new})^{-1} \big]_{\vect{\Theta} \vect{\Theta}} $}
       
   
       \If {$\frac{1}{2} \textrm{log} \frac{ \big| \vect{\Sigma}_{\vect{\Theta}|\mathcal{D}^{info}} \big| }{ \big| \vect{\Sigma}_{\vect{\Theta}|\mathcal{D}^{info},\mathcal{D}^{new}} \big| } > \lambda$ \textrm{or} $\mathcal{D}^{info} == \varnothing$}
       \State {$\mathcal{D}^{info} \gets \mathcal{D}^{info} \cup \mathcal{D}^{new}$ \, (Keep $\vect{J}(\mathcal{D}^{info})|_{\vect{\theta}^{0}}^T \vect{J}(\mathcal{D}^{info})|_{\vect{\theta}^{0}}$ as well)}
       \State {$ | \vect{\Sigma}_{\vect{\Theta}|\mathcal{D}^{info}} | \gets | \vect{\Sigma}_{\vect{\Theta}|\mathcal{D}^{info},\mathcal{D}^{new}} | $}

       \EndIf

       \EndFor

       \State {$\vect{\theta}^{\ast} \gets \texttt{Calibrate}(\mathcal{D}^{info}; \vect{\theta}^{0})$}
       \State {\textbf{return} $\vect{\theta}^{\ast}$}
   \end{algorithmic}
   \end{algorithm}
   \end{minipage}
\end{figure*}

In this section, we first present the \ref{eq:Greedy} by Maye \textit{et al.}~\cite{maye2013self,maye2016online} (Alg.~\ref{alg:maye}) 
\textcolor{black}{
\emph{as it is, without modifications.}
} 
Following that, we introduce our adaptation, which evaluates utility at initial calibration parameters (Alg.~\ref{alg:ours}). 
Additionally, we provide a theoretical analysis showing its computational complexity reduction, expressed in Big O notation.

\subsection{Original Greedy Algorithm}

\textcolor{black}{
We provide details of Alg.~\ref{alg:maye} here.
Upon receiving new measurements $\mathcal{D}^{new}$, the algorithm obtains the calibration parameter estimate $\vect{\widehat{\theta}}^{+}$ by minimizing the sum of squared residuals associated with both $\mathcal{D}^{info}$ and $\mathcal{D}^{new}$, starting from an initial estimate $\vect{\widehat{\theta}}^{-}$ (Line 6).
Next, the Jacobian --- stacks of derivatives of the residuals (see Section~\ref{subsection:Intra-Segment Residuals}) with respect to the estimated parameters $\vect{\Theta}$ --- is evaluated. 
The residuals corresponding to both $\mathcal{D}^{info}$ and $\mathcal{D}^{new}$, along with the current estimate $\vect{\widehat{\theta}}^{+}$, are used for evaluating the Jacobian.
Then, the Fisher information matrix $\vect{\mathcal{I}}$ is computed (Line 7), and the block entries corresponding to $\vect{\Theta}$ from the inverse of $\vect{\mathcal{I}}$ are taken (Line 8).
If the utility of $\mathcal{D}^{new}$, when added to $\mathcal{D}^{info}$, exceeds $\lambda$, then $\mathcal{D}^{new}$ is considered a high-utility segment (Line 9), prompting parameter updates (Lines 10-12). Note that this process iteratively refines calibration.
}

\subsection{Greedy Algorithm with Utility Evaluation at Initial Calibration Parameters}

\textcolor{black}{
Alg.~\ref{alg:ours} distinguishes itself from Alg.~\ref{alg:maye} by eliminating parameter recalibration within its loop, which, along with Fisher information matrix evaluation, is one of the primary bottlenecks in Alg.~\ref{alg:maye}. 
This change was motivated by the need to improve efficiency without compromising the results.
}
It is based on the hypothesis that utility evaluation for measurements is unaffected by the specific choice of $\vect{\theta}$, thus not significantly altering the choice of $\mathcal{D}^{info}$ or the calibration results.

\textcolor{black}{
Upon receiving new measurements $\mathcal{D}^{new}$, Alg.~\ref{alg:ours} evaluates the Jacobian \emph{only} for $\mathcal{D}^{new}$ using uncalibrated parameters $\vect{\theta}^0$, computing the Fisher information matrix $\vect{\mathcal{I}}$ (Line 5). 
The repetitive calculation of $\vect{J}(\mathcal{D}^{info})|_{\vect{\theta}^{0}}^T \vect{J}(\mathcal{D}^{info})|_{\vect{\theta}^{0}}$ is avoided by reusing results from the previous iteration (Line 8).
Then, the block entries corresponding to $\vect{\Theta}$ from the inverse of $\vect{\mathcal{I}}$ are taken (Line 6).
If the utility of $\mathcal{D}^{new}$, when added to $\mathcal{D}^{info}$, exceeds $\lambda$, then $\mathcal{D}^{new}$ is considered a high-utility segment (Line 7), prompting parameter updates and retention of $\vect{J}(\mathcal{D}^{info})|_{\vect{\theta}^{0}}^T \vect{J}(\mathcal{D}^{info})|_{\vect{\theta}^{0}}$ for future iterations (Lines 8-9).
Finally, with $\mathcal{D}^{info}$ identified, the calibration is conducted by solving a nonlinear least-squares problem (Line 12).
}

\subsection{Comparison of Time Complexity}

Consequently, Alg.~\ref{alg:ours} yields reduced time complexity compared to Alg.~\ref{alg:maye}, as shown in Table~\ref{table:big O}. Detailed derivations are provided in Appendices \ref{appendix:maye} and \ref{appendix:ours}, respectively.

Alg.~\ref{alg:ours} is notably effective when the high-utility subset selection $\mathcal{D}^{info}$ is not highly sensitive to variations in $\vect{\theta}$. To validate this, we conduct a sensitivity analysis on the impact of $\vect{\theta}$ deviations in measured information both in simulations (Section~\ref{subsection:Sensitivity Analysis in Simulation}) and hardware experiments (Section~\ref{subsection:Sensitivity Analysis on Hardware}). 
We also show that calibration results between Alg.~\ref{alg:maye} and Alg.~\ref{alg:ours} are comparably insignificant in both the simulation (Section~\ref{section:Simulation}) and hardware experiment sections (Section~\ref{section:Hardware}).

\renewcommand{\arraystretch}{1.0}
\begin{table}[h]
   \captionsetup{justification=centering, skip=0pt}
   \caption{\textsc{Big O Comparison: Original Greedy vs. Greedy with Utility Evaluation at Initial Calibration Parameters}}
   \label{table:big O}
   \begin{center}
   \begin{tabular}{|c|cc|}
       \Xhline{\arrayrulewidth}
        & Original (Alg.~\ref{alg:maye}) & Init-Param (Alg.~\ref{alg:ours}) \\
       \Xhline{\arrayrulewidth}
       Best & $\mathcal{O}(L K p^2)$ & $\mathcal{O}(L K p^2)$ \\
       Worst & $\mathcal{O}(L^2 K p^2)$ & $\mathcal{O}(L K p^2)$ \\
       \Xhline{\arrayrulewidth}
   \end{tabular}
   \end{center}
   \captionsetup{justification=raggedright}
   \caption*{\footnotesize \textit{* $L$: the number of total segments; $K$: the number of timesteps (i.e., data) in each segment; $p$: the number of parameters being estimated}}
\end{table}
\renewcommand{\arraystretch}{1.0}

\section{Self-Calibration for Multiple IMUs}
\label{section:Methodology}

This section presents the solution approach for multi-IMU extrinsic self-calibration with sparse measurement subsets, beginning with a problem overview from Section~\ref{subsection:Notation} to Section~\ref{subsection:Problem statement}, which serves as a review of our previous work~\cite{lee2022extrinsic}.
We detail a method for estimating each IMU's relative orientation in Section~\ref{subsection:Relative Orientation Initialization}, crucial for Alg.~\ref{alg:ours}'s validity, as shown in Section~\ref{subsection:Sensitivity Analysis in Simulation} and \ref{subsection:Sensitivity Analysis on Hardware}. Section~\ref{subsection:Intra-Segment Residuals} discusses residuals for evaluating the Fisher information matrix and formulating the calibration's nonlinear least-squares problem. Section~\ref{subsection:Solution Approach} introduces additional residuals for constraining the relationship between the sparse measurement subsets, followed by the calibration solution approach.

\subsection{Notation}
\label{subsection:Notation}

We denote the rotation matrix that represents frame $\mathcal{F}_A$'s orientation in frame $\mathcal{F}_B$'s coordinates by ${}^{B}_{A}\vect{R}\:\in\:SO(3)$.
The operator $\vect{C}(\cdot)$ converts a unit quaternion ${}^{B}_{A}\vect{q}$ to its corresponding rotation matrix ${}^{B}_{A}\vect{R}$.
Coordinate transformations are thus defined:
\[
    {}^{B}\vect{p} = {}^{B}_{A}\vect{R} {}^{A}\vect{p} = \vect{C}\big({}^{B}_{A}\vect{q}\big){}^{A}\vect{p}.
\]
We distinguish quantities using these symbols:
\begin{itemize}
    \setlength\itemsep{0.2em}
    \item a tilde $\widetilde{(\,\cdot\,)}$ denotes a sensor measurement,
    \item a hat $\widehat{(\,\cdot\,)}$ denotes an estimate,
    \item a zero superscript $(\,\cdot\,)^0$ denotes an initial guess (pre-optimization),
    \item and an asterisk superscript $(\,\cdot\,)^{*}$ denotes a final value (post-optimization).
\end{itemize}

\subsection{Inertial Sensor Model}

We model accelerometer measurements from each IMU as
\begin{equation}
\label{eq:accl_measurement}
    ^{I}\widetilde{\vect{a}} = {}^{I}\vect{a}_{WI} - {}^{I}\vect{g} + \vect{b}_{a} + \vect{n}_{a}
\end{equation}
where
\begin{itemize}
\item ${}^{I}\vect{a}_{WI}$ is the IMU's linear acceleration relative to the world frame, expressed in the IMU frame,
\item ${}^{I}\vect{g}$ is gravitational acceleration in the IMU frame,
\item $\vect{b}_{a}$ is a time-varying bias modeled by a random walk
\[
\vect{b}_{a, k+1} - \vect{b}_{a, k} \sim \sigma_{\vect{b}_a} \sqrt{\Delta t} \cdot \mathcal{N}\left(\vect{0}, \, \vect{1} \right),
\]
\item and $\vect{n}_{a}$ is stochastic noise
\[
\vect{n}_{a} \sim \sigma_{a} / \sqrt{\Delta t} \cdot \mathcal{N}\left(\vect{0}, \, \vect{1} \right)
\]
\end{itemize}
with $\Delta t$ as the IMU's sampling interval.
We model gyroscope measurements from each IMU as 
\begin{equation}
\label{eq:gyro_measurement}
    ^{g}\widetilde{\vect{\omega}} = \vect{C}\big({}^{g}_{I}\vect{q}\big) {}^{I}\vect{\omega}_{WI} + \vect{b}_{g} + \vect{n}_{g}
\end{equation}
where
\begin{itemize}
\item $\vect{C}(^{g}_{I}\vect{q})$ is the rotation matrix (written in terms of the corresponding quaternion) that models the IMU frame's orientation relative to the gyroscope frame, accounting for gyroscope misalignment,
\item ${}^{I}\vect{\omega}_{WI}$ is the IMU's angular velocity relative to the world frame, expressed in the IMU frame,
\item $\vect{b}_{g}$ is a time-varying bias modeled a random walk
\[
\vect{b}_{g, k+1} - \vect{b}_{g, k} \sim \sigma_{\vect{b}_g} \sqrt{\Delta t} \cdot \mathcal{N}\left(\vect{0}, \, \vect{1} \right),
\]
\item and $\vect{n}_{g}$ is stochastic noise
\[
\vect{n}_{g} \sim \sigma_{g} / \sqrt{\Delta t} \cdot \mathcal{N}\left(\vect{0}, \, \vect{1} \right).
\]
\end{itemize}
We assume that noise densities $\sigma_{a}$, $\sigma_{g}$ and bias instabilities $\sigma_{\vect{b}_a}$, $\sigma_{\vect{b}_g}$ in continuous-time are identified for all sensors before extrinsic calibration. 
Additionally, we assume intrinsic parameters like scale factors and axis non-orthogonality are calibrated, and all IMUs are temporally synchronized.


\subsection{Problem Statement (Multi-IMU Extrinsic Calibration)}
\label{subsection:Problem statement}

Supposing $N+1$ IMUs whose frames are indexed as 
$I_{0}, \, \dotsc, \, I_{N}$, 
we define:
\begin{itemize}
    \item ${}^{I_n}\vect{p}_{I_n I_0} := \vect{p}_{{I}_n}$ for the frame $I_{0}$'s position relative to frame $I_{n}$, expressed in frame $I_{n}$,
    \item ${}^{I_n}_{I_0}\vect{q} := \vect{q}_{I_n}$ for the quaternion describing frame $I_0$'s orientation relative to frame $I_n$, assuming IMU frames are accelerometer-aligned,
    \item and ${}^{I_n}_{g_n}\vect{q} := \vect{q}_{g_n}$ for the quaternion describing frame $g_n$'s orientation relative to frame $I_n$ (gyroscope misalignment).
\end{itemize}
Our calibration aims to estimate extrinsic parameters $\vect{p}_{{I}_n}$ and $\vect{q}_{{I}_n}$ for each IMU $n \in \{ 1, \dotsc, N \}$ and $\vect{q}_{g_n}$ for each IMU $n \in \{ 0, \dotsc, N \}$,
given measurements
$ ^{I_n}\widetilde{\vect{a}}_{k}, \, ^{g_n}\widetilde{\vect{\omega}}_{k} $
at each time $k$ for each IMU $n \in \{ 0, \dotsc, N \}$.
Additionally, estimating time-varying biases
$ \vect{b}_{a_{n}, k}, \, \vect{b}_{g_{n}, k} $
at each time $k$ for each IMU $n \in \{ 0, \dotsc, N \}$, as well as the angular acceleration of the base IMU
$ ^{I_0}\vect{\alpha}_{I_0, k} := ^{I_0}\vect{\alpha}_{k} $
at each time $k$, proves useful in refining sensor parameter estimates in our previous study~\cite{lee2022extrinsic}.

\subsection{Relative Orientation Initialization}
\label{subsection:Relative Orientation Initialization}

Prior to subset selection in Alg.~\ref{alg:ours}, we initialize each IMU's relative orientation to the base IMU, ${}^{I_0}_{I_n}\vect{q}$, using Yang \textit{et al.}'s method~\cite{yang2016monocular}.
This preliminary step enhances Alg.~\ref{alg:ours}'s validity by providing plausible IMU orientation guesses.

Assuming angular velocity $\vect{\omega}$ remains constant between $t_{k}$ and $t_{k+1}$, with $\Delta t = t_{k+1} - t_{k}$, the orientation change in the world frame expressed in the IMU frame is
\[
{}^{I}_{W}\vect{q}(t_{k+1}) =
\begin{bmatrix}
\frac{\vect{\omega}}{|\vect{\omega}|} \sin \left( \frac{|\vect{\omega}|}{2} \Delta t \right) \\
\cos \left( \frac{|\vect{\omega}|}{2} \Delta t \right)
\end{bmatrix}
\; \otimes \; {}^{I}_{W}\vect{q}(t_{k}).
\]
Hence, the rotation from $\mathcal{F}_{I(t_{k})}$ to $\mathcal{F}_{I(t_{k+1})}$ is
\[
{}^{I(t_{k+1})}_{I(t_{k})}\vect{q} = 
\begin{bmatrix}
\frac{\vect{\omega}}{|\vect{\omega}|} \sin \left( \frac{|\vect{\omega}|}{2} \Delta t \right) \\
\cos \left( \frac{|\vect{\omega}|}{2} \Delta t \right)
\end{bmatrix}.
\]
Supposing that we have $^{I_{0}(t_{k+1})}_{I_{0}(t_{k})}\vect{q}$ and $^{I_{n}(t_{k+1})}_{I_{n}(t_{k})}\vect{q}$ from $^{I_0}\widetilde{\vect{\omega}}_{k}$ and $^{I_n}\widetilde{\vect{\omega}}_{k}$ respectively, the following relationship holds:
\[
^{I_{0}(t_{k+1})}_{I_{0}(t_{k})}\vect{q} \, \otimes \, ^{I_0}_{I_n}\vect{q} \; = \; ^{I_0}_{I_n}\vect{q} \, \otimes \, ^{I_{n}(t_{k+1})}_{I_{n}(t_{k})}\vect{q}.
\]
Or equivalently,
\[
\mathcal{L}(^{I_{0}(t_{k+1})}_{I_{0}(t_{k})}\vect{q}) \, \cdot \, ^{I_0}_{I_n}\vect{q} = \mathcal{R}(^{I_{n}(t_{k+1})}_{I_{n}(t_{k})}\vect{q}) \, \cdot \, ^{I_0}_{I_n}\vect{q}
\]
where
\begin{align*}
\mathcal{L}(\vect{q}) &= 
\begin{bmatrix}
q_w \vect{I}_{3} + \lfloor \vect{q}_{xyz} \rfloor & \vect{q}_{xyz} \\
-\vect{q}_{xyz}^{T} & q_w
\end{bmatrix}, \\
\mathcal{R}(\vect{q}) &= 
\begin{bmatrix}
q_w \vect{I}_{3} - \lfloor \vect{q}_{xyz} \rfloor & \vect{q}_{xyz} \\
-\vect{q}_{xyz}^{T} & q_w
\end{bmatrix}.
\end{align*}
Rearranging the equation yields
\[
\left( \mathcal{L}(^{I_{0}(t_{k+1})}_{I_{0}(t_{k})}\vect{q}) - \mathcal{R}(^{I_{n}(t_{k+1})}_{I_{n}(t_{k})}\vect{q}) \right) \, \cdot \, ^{I_0}_{I_n}\vect{q} = \vect{0}_{4 \times 1}.
\]
Stacking the difference of 4-by-4 matrices $\mathcal{L} - \mathcal{R}$ for various timesteps $k$ creates a large matrix $\vect{A} \in \mathbb{R}^{4k \times 4}$, leading to $\vect{A} \, \cdot \, ^{I_0}_{I_n}\vect{q} = \vect{0}_{4 \times 1}$. We compute $^{I_0}_{I_n}\vect{q}$ by finding the right-singular vector of $\vect{A}$ with the smallest singular value.

\subsection{Intra-Segment Residuals}
\label{subsection:Intra-Segment Residuals}

For each segment $\mathcal{D}^{new}$ with $N+1$ IMU measurements over $K$ timesteps, we define four types of residuals, $\vect{r}_a$, $\vect{r}_g$, $\vect{r}_{\vect{b}_a}$, and $\vect{r}_{\vect{b}_g}$. Residual $\vect{r}_a$ relates the $n$th accelerometer's measurement ($^{I_n}\widetilde{\vect{a}}_{k}$) to that of base accelerometer's ($^{I_0}\widetilde{\vect{a}}_{k}$) corrected for bias $\vect{b}_{a_0, k}$:
\begin{equation*}
    \vect{r}_{a} = {}^{I_0}\widehat{\vect{a}}_{k} - ({}^{I_0}\widetilde{\vect{a}}_{k} - \vect{b}_{a_0, k}).
\end{equation*}
$^{I_0}\widehat{\vect{a}}_{k}$ is derived by transforming $^{I_n}\widetilde{\vect{a}}_{k}$ to the base IMU's frame:
\begin{align*}
    {}^{I_0}\widehat{\vect{a}}_{k} = \; &{}^{I_n}_{I_0}\vect{R}^{-1} \Big\{ ({}^{I_n}\widetilde{\vect{a}}_{k} - \vect{b}_{a_n, k}) + \\ & \lfloor {}^{I_n}_{g_n}\vect{R} (^{g_n}\widetilde{\vect{\omega}}_{k} - \vect{b}_{g_n, k}) \rfloor^2 \, \vect{p}_{I_n} + \lfloor {}^{I_n}_{I_0}\vect{R} ^{I_0}{\vect{\alpha}}_{k} \rfloor \, \vect{p}_{I_n} \Big\},
\end{align*}
where ${}^{I_n}_{I_0}\vect{R} = \vect{C}\big(\vect{q}_{I_n}\big)$, ${}^{I_n}_{g_n}\vect{R} = \vect{C}\big(\vect{q}_{g_n}\big)$, and $\lfloor \cdot \rfloor$ mapping a cross product into skew-symmetric matrix form. Residual $\vect{r}_g$ relates the $n$th gyroscope's measurement ($^{I_n}\widetilde{\vect{\omega}}_{k}$) to that of base gyroscope's ($^{I_0}\widetilde{\vect{\omega}}_{k}$) corrected for biases $\vect{b}_{g_n, k}$, $\vect{b}_{g_0, k}$:
\begin{equation*}
    \vect{r}_{g} = {}^{I_n}_{I_0}\vect{R}^{-1} {}^{I_n}_{g_n}\vect{R} ({}^{g_n}\widetilde{\vect{\omega}}_{k} - \vect{b}_{g_n, k}) -  {}^{I_0}_{g_0}\vect{R} ({}^{g_0}\widetilde{\vect{\omega}}_{k} - \vect{b}_{g_0, k}),
\end{equation*}
where ${}^{I_n}_{I_0}\vect{R} = \vect{C}\big(\vect{q}_{I_n}\big)$, ${}^{I_n}_{g_n}\vect{R} = \vect{C}\big(\vect{q}_{g_n}\big)$, and ${}^{I_0}_{g_0}\vect{R} = \vect{C}\big(\vect{q}_{g_0}\big)$. This relation holds as rotational motion is uniform across all points of a rigid body. 
Sensor measurements corrected for bias introduce uncertainty due to zero-mean Gaussian white noise, as shown by $\widetilde{\vect{a}} - \vect{b}_a = \vect{n}_a$ from Eq.~(\ref{eq:accl_measurement}) and $\widetilde{\vect{\omega}} - \vect{b}_g = \vect{n}_g$ from Eq.~(\ref{eq:gyro_measurement}), leading to accelerometer and gyroscope covariance matrices $ \vect{\Sigma}_{\widetilde{\vect{a}}} = ({\sigma_{a}}^2 / \Delta t) \cdot \vect{I}_{3 \times 3} $ and $ \vect{\Sigma}_{\widetilde{\vect{\omega}}} = ({\sigma_{g}}^2 / \Delta t) \cdot \vect{I}_{3 \times 3} $, respectively. 
Additional uncertainties from $^{I_0}\widetilde{\vect{\omega}}_{k}^2$ and $^{I_0}{\vect{\alpha}}_{k}$ introduced by $^{I_0}\widehat{\vect{a}}_{k}$ lead to covariance matrices $\vect{\Sigma}_{\widetilde{\vect{\omega}}^2} = 2 \cdot \vect{\Sigma}_{\widetilde{\vect{\omega}}}^2$ and $\vect{\Sigma}_{\vect{\alpha}} = \sigma_{\vect{\alpha}}^2 \cdot \vect{I}_{3 \times 3}$, respectively.
Consequently, the covariance matrix for $\vect{r}_a$ is $\vect{\Sigma}_{\vect{r}_a} = 2 \cdot \vect{\Sigma}_{\widetilde{\vect{a}}} + 2 \cdot \vect{\Sigma}_{\widetilde{\vect{\omega}}}^2 + \vect{\Sigma}_{\vect{\alpha}}$ and that for $\vect{r}_g$ is $\vect{r}_g = 2 \cdot \vect{\Sigma}_{\widetilde{\vect{\omega}}}$.
Residuals $\vect{r}_{\vect{b}_a}$ and $\vect{r}_{\vect{b}_g}$ account for bias evolution:
\begin{align*}
    \vect{r}_{\vect{b}_a} &= \vect{b}_{a_n,k+1} - \vect{b}_{a_n,k} \\
    \vect{r}_{\vect{b}_g} &= \vect{b}_{g_n,k+1} - \vect{b}_{g_n,k}.
\end{align*}
The covariance matrices for these residuals are $\vect{\Sigma}_{\vect{b}_a} = {\sigma_{\vect{b}_a}}^2 \Delta t \cdot \vect{I}_{3 \times 3}$ and $\vect{\Sigma}_{\vect{b}_g} = {\sigma_{\vect{b}_g}}^2 \Delta t \cdot \vect{I}_{3 \times 3}$, respectively.


\subsection{Solution Approach}
\label{subsection:Solution Approach}

Upon identifying high-utility measurements, we formulate a nonlinear least-square problem using the chosen measurement subset. Beyond the previously mentioned residuals in Section~\ref{subsection:Intra-Segment Residuals}, we incorporate additional residuals, $\vect{r}_{\vect{b}_a}^{\ast}$ and $\vect{r}_{\vect{b}_g}^{\ast}$, to bridge temporal gaps between adjacent selected segments, denoted as $\mathcal{D}^{info} = \{ \mathcal{D}^m | m \in \mathcal{M} \}$ with $\mathcal{M} = \{ \mathcal{M}_1, \dotsc, \mathcal{M}_M  \} \subseteq \{1, \dotsc, L \}$:
\begin{align*}
    \vect{r}_{\vect{b}_a}^{\ast} &= \vect{b}_{a_n,(\mathcal{M}_{m+1}-1) \cdot K + 1} - \vect{b}_{a_n,\mathcal{M}_{m} \cdot K} \\
    \vect{r}_{\vect{b}_g}^{\ast} &= \vect{b}_{g_n,(\mathcal{M}_{m+1}-1) \cdot K + 1} - \vect{b}_{g_n,\mathcal{M}_{m} \cdot K}.
\end{align*}
These additional residuals aim to maintain bias consistency across segments, with their covariance matrices $\vect{\Sigma}_{\vect{b}_a}^{\ast}$ and $\vect{\Sigma}_{\vect{b}_g}^{\ast}$ modeled after a random walk in discrete time as earlier discussed, resulting in $\vect{\Sigma}_{\vect{b}_a}^{\ast} = {\sigma_{\vect{b}_a}}^2 \cdot (\mathcal{M}_{m+1}-\mathcal{M}_{m})K \cdot \Delta t \cdot \vect{I}_{3 \times 3}$ and $\vect{\Sigma}_{\vect{b}_g}^{\ast} = {\sigma_{\vect{b}_g}}^2 \cdot (\mathcal{M}_{m+1}-\mathcal{M}_{m})K \cdot \Delta t \cdot \vect{I}_{3 \times 3}$.

These residuals allow for parameter estimation through solving a nonlinear least-squares problem
\begin{equation}
    \label{eq:problem statement}
    \resizebox{\columnwidth}{!}{
    $
    \begin{split}
        \min \Bigg\{ 
        \sum_{l \in \mathcal{M}} 
        & \Big(
        \sum_{ \substack{n \in \{1, \dotsc, N\} \\ k \in \{1, \dotsc, K\} } } \big( \left\| \vect{r}_{a} \right\|^{2}_{\vect{\Sigma}_{\vect{r}_a}} + \left\| \vect{r}_{g}  \right\|^{2}_{\vect{\Sigma}_{\vect{r}_g}} \big) \\
        + &
        \sum_{ \substack{n \in \{0, \dotsc, N\} \\ k \in \{1, \dotsc, K-1\} } } \big( \left\| \vect{r}_{\vect{b}_a} \right\|^{2}_{\vect{\Sigma}_{\vect{r}_{\vect{b}_{a}}}} + \left\| \vect{r}_{\vect{b}_g} \right\|^{2}_{\vect{\Sigma}_{\vect{r}_{\vect{b}_{g}}}} \big) 
        \Big) \\
        + 
        \sum_{ \substack{m \in \{1, \dotsc, M-1 \} \\ n \in \{0, \dotsc, N\} } } &
        \big( \left\| \vect{r}_{\vect{b}_a}^{\ast} \right\|^{2}_{\vect{\Sigma}_{\vect{r}_{\vect{b}_{a}}}^{\ast}} + \left\| \vect{r}_{\vect{b}_g}^{\ast} \right\|^{2}_{\vect{\Sigma}_{\vect{r}_{\vect{b}_{g}}}^{\ast}} \big)
        \Bigg\},
    \end{split}
    $
    }
\end{equation}
where $ \left\| \cdot \right\|^{2}_{\vect{\Sigma}}$ represents the Mahalanobis distance using covariance matrix $\vect{\Sigma}$.

\newpage
\section{Simulation Experiments}
\label{section:Simulation}

In this section, we assess multi-IMU extrinsic self-calibration using simulated trajectories from OpenVINS~\cite{geneva2020openvins}.
First, in Section~\ref{subsection:Limitations of Existing Methods for Self-Calibration with Informative Subset Selection}, we present an edge case highlighting the need for adopting the \ref{eq:Greedy} for subset selection over \ref{eq:M-largest}, also noting the original \ref{eq:Greedy} (Alg.~\ref{alg:maye})'s efficiency issue.
A sensitivity analysis in Section~\ref{subsection:Sensitivity Analysis in Simulation} examines the impact of changes in parameter estimates on the measure of information---a function of the parameter estimates. This analysis supports the use of Alg.~\ref{alg:ours} as an alternative to Alg.~\ref{alg:maye}. 
\textcolor{black}{
Section~\ref{subsection:Comparison in Simulation} compares the calibration performances of Alg.~\ref{alg:maye} (\emph{Greedy (Original)}) and Alg.~\ref{alg:ours} (\emph{Greedy (Init-Param)}) against the \emph{baseline}, which uses the full set of measurements for calibration, showing Alg.~\ref{alg:ours} significantly reduces runtime---from minutes to roughly a quarter minute---without sacrificing accuracy.
}

\subsection{Implementation Details}

\renewcommand{\arraystretch}{1.0}
\begin{table}[h]
    \captionsetup{justification=centering, skip=0pt}
    \caption{\textsc{Reference IMU poses in simulation}}
    \label{table:sensor layout}
    \centering
    \footnotesize
    \begin{tabular}{|ccc|}
        \Xhline{\arrayrulewidth}
        index & position [cm] & orientation [deg]* \\
        \Xhline{\arrayrulewidth}
        IMU0  & [0,  0, 0] & [0, 0, 0] \\
        IMU1  & [20, 0, 0] & [180, 0, 0] \\
        IMU2  & [0, 20, 0] & [0, 180, 0] \\
        IMU3  & [0, 0, 20] & [0, 0, 180] \\
        \Xhline{\arrayrulewidth}
    \end{tabular}
    \captionsetup{justification=raggedright}
    \caption*{\footnotesize \textit{* in this table, orientation is given as XYZ Euler angles}}
\end{table}
\renewcommand{\arraystretch}{1.0}

Table~\ref{table:sensor layout} shows the reference poses for four IMUs mounted on a rigid body.
Accelerometer noise and bias instability were set at $\sigma_{a} = 2 \times 10^{-3}~\textrm{m} / \textrm{s}^{2} / \sqrt{\textrm{Hz}}$ and $\sigma_{\vect{b}_a} = 3 \times 10^{-3}~\textrm{m} / \textrm{s}^{2} \cdot \sqrt{\textrm{Hz}}$, respectively, while for gyroscopes, these values were $\sigma_{g} = 1.6968 \times 10^{-4}~\textrm{rad} / \textrm{s} / \sqrt{\textrm{Hz}}$ and $\sigma_{\vect{b}_g} = 1.9393 \times 10^{-5}~\textrm{rad} / \textrm{s} \cdot \sqrt{\textrm{Hz}}$.
Gyroscope misalignment was generated by rotating each IMU's reference orientation about a uniformly random axis by an angle from a zero-mean normal distribution with a standard deviation of $1^{\circ}$.
IMU measurements were generated at 100 Hz.

\subsection{Limitations of Existing Methods for Self-Calibration with Informative Subset Selection}
\label{subsection:Limitations of Existing Methods for Self-Calibration with Informative Subset Selection}

\begin{table*}
    \captionsetup{justification=centering}
    \caption{\textsc{\footnotesize{Error in estimated extrinsic parameters and runtime for baseline, M-largest selection, and greedy algorithm}}}
    \label{table:subset selection comparison}
    \centering
    \footnotesize
    \resizebox{0.8\linewidth}{!}{
    \begin{tabular}{|cccc|cccc|cccc|}
        \Xhline{\arrayrulewidth}
        \multicolumn{4}{|c|}{Baseline} & \multicolumn{4}{c|}{M-largest Selection} & \multicolumn{4}{c|}{Greedy Algorithm} \\
        $\vect{p}$ [cm] & $\vect{q}$ [deg] & $\vect{q}_g$ [deg] & $t$ [s] & $\vect{p}$ [cm] & $\vect{q}$ [deg] & $\vect{q}_g$ [deg] & $t$ [s] & $\vect{p}$ [cm] & $\vect{q}$ [deg] & $\vect{q}_g$ [deg] & $t$ [s] \\
        \Xhline{\arrayrulewidth}
        0.0151 & 0.3685 & 2.1002 & 7.58 & 19.8605 & 1.4150 & 62.3996 & 1.91 & 0.0229 & 0.3090 & 0.8152 & 8.04 \\
        \Xhline{\arrayrulewidth}
    \end{tabular}
    }
    \captionsetup{justification=raggedright}
\end{table*}

\begin{figure*}
    \centering
    \includegraphics[width=0.9\linewidth, keepaspectratio]{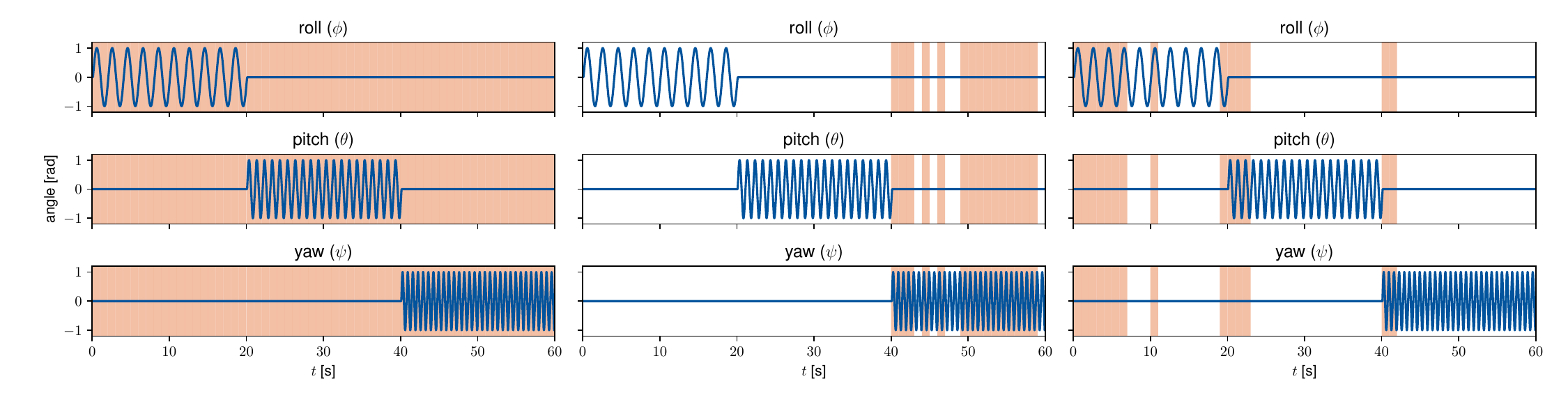}
    \caption{The intervals of subset within a roll-pitch-yaw profile, which is used by the calibration process for the baseline (left), \ref{eq:M-largest} (middle), and \ref{eq:Greedy} (right), are highlighted. \textcolor{black}{The background color highlights the segments selected and used in each calibration process.}}
    \label{fig:subset selection comparison}
\end{figure*}

We assess multi-IMU extrinsic self-calibration using three measurement subsets: the full set (``baseline''), and subsets selected by \ref{eq:M-largest} and the \ref{eq:Greedy}. 
We simulated a trajectory with sinusoidal oscillations in XYZ Euler angles every 20 seconds, maintaining consistent information for each interval:
\begin{itemize}
\item For $0 \leq t < 20$ seconds: $[\text{sin}(1 \pi t), 0, 0] ~ \textrm{[rad]}$.
\item For $20 \leq t < 40$ seconds: $[0, \text{sin}(2 \pi t), 0] ~ \textrm{[rad]}$.
\item For $40 \leq t < 60$ seconds: $[0, 0, \text{sin}(3 \pi t)] ~ \textrm{[rad]}$.
\end{itemize}
For the \ref{eq:Greedy}, the utility threshold $\lambda$ was set to 0.5, following previous studies~\cite{maye2013self,maye2016online}. For \ref{eq:M-largest}, we chose 15 segments, ensuring a similar subset size to the \ref{eq:Greedy}. Initial parameter estimates $\vect{p}_{{I}_n}$, $\vect{q}_{{I}_n}$, and $\vect{q}_{g_n}$ were set to zero, with angular acceleration derived from numerical differentiation of IMU 0's gyroscope measurements and time-varying biases starting at zero.

Table~\ref{table:subset selection comparison} presents the calibration error and runtime for each process, and Fig.~\ref{fig:subset selection comparison} illustrates the chosen segments. \ref{eq:M-largest} shows significantly higher calibration errors than both the baseline and \ref{eq:Greedy}. While \ref{eq:Greedy} selects new segments with utility (notably around 20s and 40s), the failure of \ref{eq:M-largest} in this regard likely results in poorer estimates. 
Despite lower errors, \ref{eq:Greedy}'s runtime is longer than \ref{eq:M-largest} and the baseline, highlighting the need for the efficient subset selection as introduced in Section~\ref{section:Subset Identification}.

\subsection{Sensitivity Analysis of Fisher Information Matrix}
\label{subsection:Sensitivity Analysis in Simulation}

The underlying assumption of Alg.~\ref{alg:ours} is the Fisher information matrix ($\vect{\mathcal{I}} = \vect{J}^T|_\vect{\theta} \vect{J}|_\vect{\theta}$) --- a function of the estimate for the parameters $\vect{\theta}$ --- is insensitive to the parameter estimate. To investigate this, we examined the Spearman's correlation coefficient~\cite{zar2014spearman} between the distribution of information $ - 1/2 \cdot \log \big| \big[ {\vect{\mathcal{I}}}^{-1} \big]_{\vect{\Theta}\vect{\Theta}} \big|$ at reference extrinsic parameters (\textit{original distribution}) and at values altered by deviations in either position ($\delta{\vect{p}}$) or orientation ($\delta{\vect{q}}$) (\textit{perturbed distribution}). The study utilized simulations across 48 different trajectories from five datasets provided by OpenVINS~\cite{geneva2020openvins}. 

\begin{figure}
    \centering
    \includegraphics[width=0.9\linewidth, keepaspectratio]{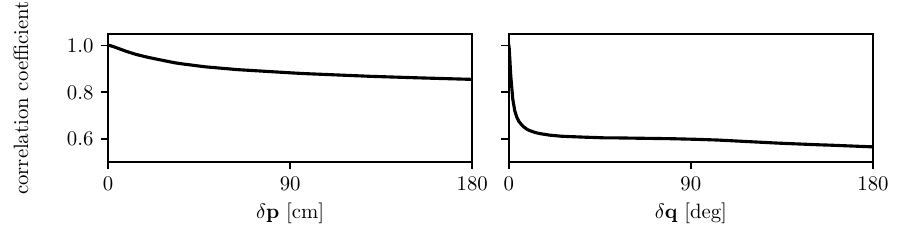}
    \caption{The correlation coefficient of the information as a function of the difference ($\delta \vect{p}$, $\delta \vect{q}$) in guessed and reference extrinsic parameters in the multi-IMU system. A coefficient of 1 indicates perfect correlation with the true value, while 0 denotes independence.}
    \label{fig:correlation}
\end{figure}

Fig.~\ref{fig:correlation} shows changes in the correlation coefficient are smaller for positional deviations ($\delta{\vect{p}}$) but larger for orientational deviations ($\delta{\vect{q}}$), even with a few degrees. This indicates information sensitivity mainly lies with orientation guesses rather than position. Hence, ensuring accurate initial orientation guesses, as detailed in Section~\ref{subsection:Relative Orientation Initialization}, is key for effective segment selection by Alg.~\ref{alg:ours}.

\subsection{Comparison of Self-Calibration with Greedy Algorithm against Benchmarks}
\label{subsection:Comparison in Simulation}

\renewcommand{\arraystretch}{1.1}
\begin{table*}
    \captionsetup{justification=centering}
    \caption{\textsc{Ratio of selected segments and runtime for baseline, original greedy, and greedy with utility evaluation at initial calibration parameters over each dataset}}
    \label{table:sparsification results}
    \centering
    \resizebox{\linewidth}{!}{
    \begin{tabular}{|cccc|ccc|ccc|}
        \Xhline{\arrayrulewidth}
        \multicolumn{4}{|c|}{} & \multicolumn{3}{c|}{Ratio of selected segments [$\%$]} & \multicolumn{3}{c|}{Runtime [s]} \\
        Dataset & $\#$ of traj. & Length [m] & Duration [s] & Baseline & Greedy (Original) & Greedy (Init-Param) & Baseline & Greedy (Original) & Greedy (Init-Param) \\
        \Xhline{\arrayrulewidth}
        \texttt{uzh\_fpv}   & 15 &  210.02 $\pm$  144.69 &   39.03 $\pm$  17.68 & 100.00 $\pm$ 0.00 & 59.30 $\pm$ 17.37 & 66.15 $\pm$ 21.01 &   5.98 $\pm$   3.46 &   7.12 $\pm$   4.44 & \textbf{ 3.41 $\pm$ 0.94} \\
        \texttt{euroc\_mav} & 11 &   81.22 $\pm$   23.44 &  122.58 $\pm$  27.58 & 100.00 $\pm$ 0.00 &  26.06 $\pm$ 8.43 & 24.88 $\pm$  7.41 &  18.21 $\pm$   4.78 &  23.16 $\pm$   6.22 & \textbf{ 5.72 $\pm$ 1.16} \\
        \texttt{tum\_vi}    &  6 &  115.25 $\pm$   36.98 &  135.10 $\pm$  12.52 & 100.00 $\pm$ 0.00 &  25.94 $\pm$ 7.03 & 26.33 $\pm$  5.11 &  19.82 $\pm$   1.99 &  32.45 $\pm$  10.06 & \textbf{ 5.72 $\pm$ 1.06} \\
        \texttt{kaist\_vio} & 11 &   37.10 $\pm$   16.54 &  179.44 $\pm$  30.00 & 100.00 $\pm$ 0.00 &  17.76 $\pm$ 9.74 & 17.66 $\pm$  6.39 &  25.28 $\pm$  11.88 &  30.91 $\pm$  19.29 & \textbf{ 5.32 $\pm$ 1.79} \\
        \texttt{kaist}      &  5 & 9618.87 $\pm$ 1958.50 & 1533.96 $\pm$ 656.45 & 100.00 $\pm$ 0.00 &   3.87 $\pm$ 2.26 &  3.44 $\pm$  2.11 & 379.42 $\pm$ 135.80 & 460.81 $\pm$ 211.07 & \textbf{20.08 $\pm$ 6.60} \\
        \Xhline{\arrayrulewidth}
    \end{tabular}
    }
\end{table*}
\renewcommand{\arraystretch}{1.0}

\renewcommand{\arraystretch}{1.1}
\begin{table*}
    \captionsetup{justification=centering}
    \caption{\textsc{Runtime breakdown for baseline, original greedy, and greedy with utility evaluation at initial calibration parameters over each dataset}}
    \label{table:runtime breakdown}
    \centering
    \resizebox{\linewidth}{!}{
    \begin{tabular}{|c|ccc|ccc|ccc|ccc|}
        \Xhline{\arrayrulewidth}
         & \multicolumn{3}{c|}{Baseline} & \multicolumn{3}{c|}{Greedy (Original)} & \multicolumn{3}{c|}{Greedy (Init-Param)} \\
        Dataset & \texttt{Evaluate} & \texttt{Calibrate} & Total & \texttt{Evaluate} & \texttt{Calibrate} & Total & \texttt{Evaluate} & \texttt{Calibrate} & Total \\
        \Xhline{\arrayrulewidth}
        \texttt{uzh\_fpv}   & - &     5.94 $\pm$ 3.43 &   5.98 $\pm$   3.46 &   2.52 $\pm$   2.05 &  4.50 $\pm$  2.39 &   7.12 $\pm$   4.44 & 0.15 $\pm$ 0.07 &  3.19 $\pm$ 0.86 &  3.41 $\pm$ 0.94 \\
        \texttt{euroc\_mav} & - &    18.09 $\pm$ 4.76 &  18.21 $\pm$   4.78 &  13.84 $\pm$   4.11 &  8.91 $\pm$  2.94 &  23.16 $\pm$   6.22 & 0.46 $\pm$ 0.13 &  5.12 $\pm$ 1.19 &  5.72 $\pm$ 1.16 \\
        \texttt{tum\_vi}    & - &    19.69 $\pm$ 1.98 &  19.82 $\pm$   1.99 &  18.50 $\pm$   6.46 & 13.48 $\pm$  3.89 &  32.45 $\pm$  10.06 & 0.54 $\pm$ 0.06 &  5.04 $\pm$ 1.06 &  5.72 $\pm$ 1.06 \\
        \texttt{kaist\_vio} & - &   25.13 $\pm$ 11.82 &  25.28 $\pm$  11.88 &  21.11 $\pm$  14.18 &  9.27 $\pm$  5.43 &  30.91 $\pm$  19.29 & 0.58 $\pm$ 0.26 &  4.56 $\pm$ 1.51 &  5.32 $\pm$ 1.79 \\
        \texttt{kaist}      & - & 377.85 $\pm$ 135.22 & 379.42 $\pm$ 135.80 & 371.62 $\pm$ 172.61 & 81.47 $\pm$ 41.68 & 460.81 $\pm$ 211.07 & 4.61 $\pm$ 1.98 & 13.46 $\pm$ 5.33 & 20.08 $\pm$ 6.60 \\
        \Xhline{\arrayrulewidth}
    \end{tabular}
    }
\end{table*}
\renewcommand{\arraystretch}{1.0}

\renewcommand{\arraystretch}{1.0}
\begin{table}
    \captionsetup{justification=centering, skip=0pt}
    \caption{\textsc{Memory footprint (RAM usage in MB) for baseline, original greedy, and greedy with utility evaluation at initial calibration parameters over each dataset}}
    \label{table:memory usage}
    \centering
    \resizebox{\linewidth}{!}{
    \begin{tabular}{|c|ccc|}
        \Xhline{\arrayrulewidth}
        Dataset & Baseline & Greedy (Original) & Greedy (Init-Param) \\
        \Xhline{\arrayrulewidth}
        \texttt{uzh\_fpv} & 92.69 $\pm$ 40.62 & 55.38 $\pm$ 14.88 & 65.13 $\pm$ 32.34 \\
        \texttt{euroc\_mav} & 235.46 $\pm$ 50.42 & 82.70 $\pm$ 14.43 & 69.10 $\pm$ 33.19 \\
        \texttt{tum\_vi} & 264.21 $\pm$ 24.86 & 92.54 $\pm$ 17.72 & 100.73 $\pm$ 38.73 \\
        \texttt{kaist\_vio} & 274.96 $\pm$ 124.70 & 80.69 $\pm$ 45.38 & 85.89 $\pm$ 47.05 \\
        \texttt{kaist} & 1678.32 $\pm$ 350.95 & 141.08 $\pm$ 11.22 & 135.75 $\pm$ 8.31 \\
        \Xhline{\arrayrulewidth}
    \end{tabular}
    }
\end{table}
\renewcommand{\arraystretch}{1.0}

\begin{figure}
   \centering
   \includegraphics[width=0.70\linewidth, keepaspectratio]{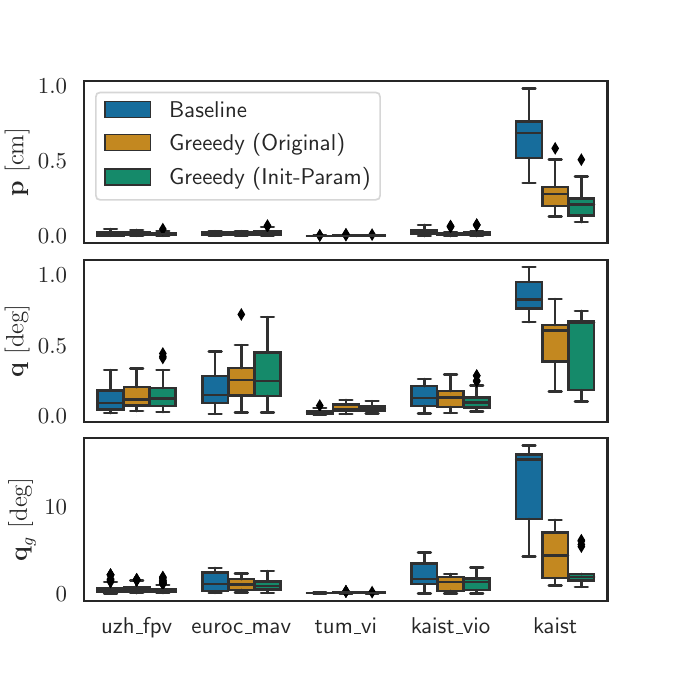}
   \caption{Absolute error in estimated extrinsic parameters for baseline, original greedy, and greedy with utility evaluation at initial calibration parameters over each dataset in simulation.}
   \label{fig:box plot}
\end{figure}

We assess the self-calibration with subset selection in Alg.~\ref{alg:ours} (``Greedy (Init-Param)'') against Alg.~\ref{alg:maye} (``Greedy (Original)'') and a ``Baseline'' using all measurements. 
Our focus lies on runtime and calibration performance comparisons across 48 trajectories within five OpenVINS-supported datasets.
For both the greedy methods, information evaluation occurred every second, using a $\lambda = 0.5$ threshold, with initial parameter guesses aligned with the earlier discussion.

Table~\ref{table:sparsification results} presents the chosen measurement subset ratio and runtime for each calibration process, with Fig.~\ref{fig:box plot} showing the absolute extrinsic parameter errors for all IMUs.
The runtime breakdown and memory usage are presented in Tables~\ref{table:runtime breakdown} and \ref{table:memory usage}, respectively.
Greedy (Init-Param) consistently reduced runtime, with significant improvements in the \texttt{kaist} dataset, while Greedy (Original) showed less or negative improvement.
Calibration accuracy remained consistent both for greedy methods and the baseline, achieving sub-centimeter and sub-degree precision, except in the \texttt{kaist} dataset. 
Greedy methods particularly outperformed the baseline in the \texttt{kaist}, which was collected in a driving car and primarily consists of low-utility measurements, by excluding these low-utility measurements to improve problem observability.  
The subset selection addresses the challenges of unobservable or weakly observable calibration scenarios that can arise in self-calibration with full measurements.

\newpage
\section{Hardware Experiments}
\label{section:Hardware}

This section evaluates self-calibration with measurement subsets using real-world data from a dataset featured in our prior work~\cite{lee2022extrinsic}. 
As in simulations, a sensitivity analysis in Section~\ref{subsection:Sensitivity Analysis on Hardware} examines how changes in parameter estimates affect the measure of information, supporting the use of Alg.~\ref{alg:ours} as an alternative to Alg.~\ref{alg:maye}.
Section~\ref{subsection:Comparison on Hardware} compares the calibration performances of Alg.~\ref{alg:maye} and Alg.~\ref{alg:ours} with a full-measurement baseline, highlighting Alg.~\ref{alg:ours}'s reduction in runtime---from approximately one minute to just a second---without loosing calibration accuracy.

\subsection{Implementation Details}

\begin{figure}[h]
	\vspace{-3.5mm} 
    \centering
    \includegraphics[width=0.32\linewidth, keepaspectratio]{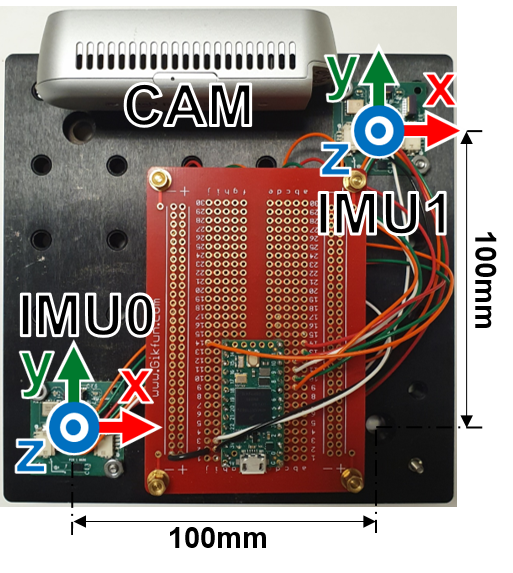}
    \caption{Sensor rig used for data collection.}
    \label{fig:sensor rig}
	\vspace{-3.5mm} 
\end{figure}

Fig.~\ref{fig:sensor rig} shows the sensor rig used for data collection, featuring two IMUs (IMU 0 and IMU 1) and a camera (CAM),  with only the IMU data utilized for evaluation.
The reference relative position $\vect{p}_{I_{1}}$ and orientation $\vect{q}_{I_{1}}$ between the IMUs are $[-10, -10, 0] \; [\textrm{cm}]$ and $[0, 0, 0] \; [\textrm{deg}]$ in XYZ Euler angles, with assumed zero gyroscope misalignment; $\vect{q}_{g_0}$ and $\vect{q}_{g_1}$ are also $[0, 0, 0] \; [\textrm{deg}]$. 
Allan variance analysis provided white noise and bias instability of the accelerometers $\sigma_{a} = 3.3 \times 10^{-3}~\textrm{m} / \textrm{s}^{2} / \sqrt{\textrm{Hz}}$ and $\sigma_{\vect{b}_a} = 2.4 \times 10^{-4}~\textrm{m} / \textrm{s}^{2} \cdot \sqrt{\textrm{Hz}}$ and of the gyroscopes $\sigma_{g} = 8.3 \times 10^{-5}~\textrm{rad} / \textrm{s} / \sqrt{\textrm{Hz}}$ and $\sigma_{\vect{b}_g} = 3.7 \times 10^{-6}~\textrm{rad} / \textrm{s} \cdot \sqrt{\textrm{Hz}}$. 

Acknowledging that comparing estimated calibration parameters to reference values might not entirely reflect accuracy, as noted in our previous work, we also measure reprojection error between the IMUs, with a lower error indicating more precise estimation.


\subsection{Sensitivity Analysis of Fisher Information Matrix}
\label{subsection:Sensitivity Analysis on Hardware}

\begin{figure}[h]
	\vspace{-1mm} 
    \centering
    \includegraphics[width=0.9\linewidth, keepaspectratio]{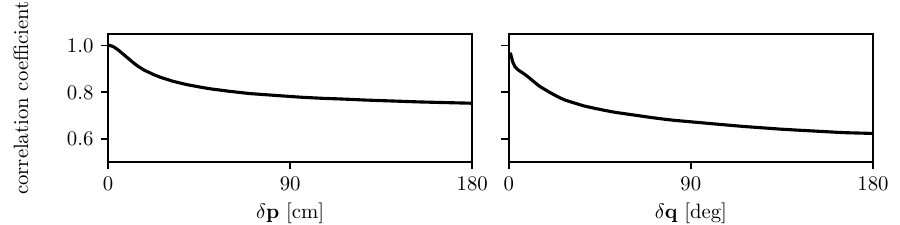}
    \caption{The correlation coefficient of the information as a function of the difference ($\delta \vect{p}$, $\delta \vect{q}$) in guessed and reference extrinsic parameters in the two-IMU system on hardware. A coefficient of 1 indicates perfect correlation with the true value, while 0 denotes independence.}
    \label{fig:correlation_hardware}
    \vspace{-1mm} 
\end{figure}

As in the sensitivity analysis conducted in simulations (in Section~\ref{subsection:Sensitivity Analysis in Simulation}), we assessed how deviations from true calibration parameters affect the measured information using the hardware data. In Fig.\ref{fig:correlation_hardware}, we examined the Spearman's correlation coefficient~\cite{zar2014spearman} between the information distribution, $ - 1/2 \cdot \log \big| \big[ {\vect{\mathcal{I}}}^{-1} \big]_{\vect{\Theta}\vect{\Theta}} \big|$, evaluated at reference extrinsic parameters and at deviations in either relative position ($\delta{\vect{p}}$) or orientation ($\delta{\vect{q}}$).

Similar to that in simulations, sensitivity to orientation deviations ($\delta{\vect{q}}$) was evident, showing a greater decline than for position deviations ($\delta{\vect{p}}$). This highlights the importance of good initial guesses for relative orientation, as advocated in Section~\ref{subsection:Relative Orientation Initialization}, for sucessful measurement subset selection.

\subsection{Comparison of Self-Calibration with Efficient Greedy Algorithm against Benchmarks}
\label{subsection:Comparison on Hardware}


\renewcommand{\arraystretch}{1.05}
\begin{table}[h]
	\vspace{-1mm} 
    \captionsetup{justification=centering, skip=0pt}
    \caption{\textsc{Estimated extrinsics, segment selection ratio, and runtime for baseline, original greedy, and greedy with utility evaluation at initial calibration parameters on hardware data}}
    \label{table:hardware-long}
    \centering
    \resizebox{\linewidth}{!}{
    \begin{tabular}{|cc|c|c|c|}
        \Xhline{\arrayrulewidth}
        Trajectory &  & Baseline & Greedy (Original) & Greedy (Init-Param) \\
        \Xhline{\arrayrulewidth}
        \multirow{6}{*}{\thead{\texttt{baseline}\\(1274 [s])}} 
         & $\vect{p}_{I_1}$ [cm] & [-9.80  -9.82 -0.24] & [-9.85 -9.63 -0.34] & [-9.64 -9.58 -0.29] \\
         & $\vect{q}_{I_1}$ [deg] & [4.15 3.28 0.27] & [ 3.39  2.01 -0.42] & [ 1.98  0.61 -0.40] \\
         & $\vect{q}_{g_0}$ [deg] & [-0.74 -0.45  0.04] & [ 2.47  4.34 -1.08] & [-0.90 1.06 0.21] \\
         & $\vect{q}_{g_1}$ [deg] & [3.11 3.31 0.45] & [ 5.62  6.82 -1.05] & [0.84 2.12 0.09] \\
         & Selected segments [\%] & 100.00 & 0.73 & 1.14 \\
         & Runtime [s] & 64.09 & 16.15 & \textbf{0.98} \\
        \Xhline{\arrayrulewidth}
        \multirow{6}{*}{\thead{\texttt{blurry}\\(1388 [s])}} 
         & $\vect{p}_{I_1}$ [cm] & [-9.81 -9.77 -0.26] & [-9.84 -9.78 -0.30] & [-9.79 -9.72 -0.27] \\
         & $\vect{q}_{I_1}$ [deg] & [5.27 4.27 0.25] & [ 2.22  0.98 -0.25] & [ 1.96  0.96 -0.65] \\
         & $\vect{q}_{g_0}$ [deg] & [-2.96 -2.74 0.00] & [-0.43 -0.06 -0.27] & [-1.76 -1.13  0.45] \\
         & $\vect{q}_{g_1}$ [deg] & [2.03 2.11 0.20] & [ 1.65  1.28 -0.19] & [0.09 0.15 0.05] \\
         & Selected segments [\%] & 100.00 & 1.94 & 0.82 \\
         & Runtime [s] & 75.81 & 51.01 & \textbf{0.90} \\
        \Xhline{\arrayrulewidth}
        \multirow{6}{*}{\thead{\texttt{ill-lit}\\(1276 [s])}} 
         & $\vect{p}_{I_1}$ [cm] & [-9.82 -9.86 -0.21] & [-9.67 -9.96 -0.14] & [-9.65 -9.83 -0.17] \\
         & $\vect{q}_{I_1}$ [deg] & [-5.03 -5.79  0.28] & [ 2.50 1.29 -0.02] & [5.03 3.50 0.10] \\
         & $\vect{q}_{g_0}$ [deg] & [-0.36 -0.34  0.02] & [ 1.70 0.69 -0.18] & [ 2.55  1.89 -0.40] \\
         & $\vect{q}_{g_1}$ [deg] & [-5.79 -5.58  0.39] & [3.82 2.42 0.02] & [7.24 6.20 0.21] \\
         & Selected segments [\%] & 100.00 & 2.19 & 1.05 \\
         & Runtime [s] & 73.75 & 59.37 & \textbf{0.99} \\
        \Xhline{\arrayrulewidth}
    \end{tabular}
    }
    \vspace{-1mm} 
\end{table}
\renewcommand{\arraystretch}{1.0}

\renewcommand{\arraystretch}{1.0}
\begin{table}[h!]
	\vspace{-1mm} 
    \captionsetup{justification=centering, skip=0pt}
    \caption{\textsc{IMU measurement reprojection errors for baseline, original greedy, and greedy with utility evaluation at initial calibration parameters on hardware data}}
    \label{table:IMU-reprojection-long}
    \centering
    \resizebox{\linewidth}{!}{
    \begin{tabular}{|cc|c|c|c|}
        \Xhline{\arrayrulewidth}
         & Dataset & Baseline & Greedy (Original) & Greedy (Init-Param) \\
        \Xhline{\arrayrulewidth}
        \multirow{3}{*}{\thead{accelerometer\\([$\textrm{m}/\textrm{s}^2$])}}
         & \texttt{baseline} & 0.8160 $\pm$ 0.4947 & \textbf{0.6015 $\pm$ 0.4947} & \textbf{0.4749 $\pm$ 0.4760} \\
         & \texttt{blurry} & 1.0981 $\pm$ 0.5470 & 0.5069 $\pm$ 0.5470 & 0.5041 $\pm$ 0.4257 \\
         & \texttt{ill-lit} & 1.6424 $\pm$ 0.3951 & \textbf{0.4697 $\pm$ 0.3951} & \textbf{0.8968 $\pm$ 0.3911} \\
        \Xhline{\arrayrulewidth}
        \multirow{3}{*}{\thead{gyroscope\\([$\textrm{rad}/\textrm{s}$])}}
         & \texttt{baseline} & 0.0942 $\pm$ 0.0139 & 0.0942 $\pm$ 0.0139 & 0.0942 $\pm$ 0.0141 \\
         & \texttt{blurry} & 0.0996 $\pm$ 0.0209 & 0.0996 $\pm$ 0.0219 & 0.0996 $\pm$ 0.0219 \\
         & \texttt{ill-lit} & 0.0926 $\pm$ 0.0101 & 0.0925 $\pm$ 0.0103 & 0.0930 $\pm$ 0.0107 \\
        \Xhline{\arrayrulewidth}
    \end{tabular}
    }
	\vspace{-1mm} 
\end{table}
\renewcommand{\arraystretch}{1.0}

Table~\ref{table:hardware-long} shows estimated calibration parameters $\vect{p}_{I_1}$, $\vect{q}_{I_1}$, $\vect{q}_{g_0}$, and $\vect{q}_{g_1}$ along with the chosen measurement subset ratio and runtime for each calibration process. 
Table~\ref{table:IMU-reprojection-long} shows the IMU reprojection error across trajectories for the three methods.
\textcolor{black}{
Significant differences between the greedy methods, determined through t-tests, are highlighted in bold.
}
Initial parameter guesses were as previously discussed in simulations.
The results indicate the greedy algorithms use less than 3\% of total measurements to achieve calibration results that closely align with reference values. Greedy (Init-Param) significantly reduces runtime from over a minute to approximately a second, whereas Greedy (Original) achieves a modest reduction, cutting the runtime by about fifteen seconds from the baseline.

\section{Conclusion}
\label{section:Conclusion}

This paper presents a multi-IMU extrinsic calibration by selecting high-utility measurement subsets. We hypothesize that, in our system, utility---a function of the parameter estimates---is largely insensitive to the specific choice of parameters, allowing evaluation at an initial guess and reducing the need for frequent recalibrations. This significantly decreases computation time in both simulations and experiments, making it suitable for resource-limited platforms.

\textcolor{black}{
We acknowledge that the hypothesis regarding the insensitivity in parameter choice for subset selection could have been examined more thoroughly.
Future research could explore this in greater depth, for example by comparing the selected segments using both the original and the modified Greedy approaches.
}


\appendices

\section{Proof of Time Complexity of Original Greedy Algorithm}
\label{appendix:maye}

In this appendix, we derive the time complexity of the original greedy algorithm (Alg.~\ref{alg:maye}).

We will use the following symbols:
\begin{itemize}
    \item $L$ means the number of total segments.
    \item $M$ means the number of finally chosen segments (i.e., $M \leq L$).
    \item $K$ means the number of timesteps (i.e., data) in each segment.
    \item $p$ means the number of parameters being estimated.
\end{itemize}
We assume $K > p$ in general.

\begin{enumerate}
    \item Line 5 in Alg.~\ref{alg:maye} iterates $L$ times.
    \item Line 6 conducts nonlinear least-square on $m \times K$ data, where $m$ is the number of segments in $\mathcal{D}^{info} \cup \mathcal{D}^{new}$ ($\mathcal{O}(mK \cdot p^2)$).
    \item Line 7 evaluates Jacobians $\vect{J}(\mathcal{D}^{info})_{\vect{\hat{\theta}}^{+}}$ (size: $(m-1)K \times p$) and $\vect{J}(\mathcal{D}^{new})_{\vect{\hat{\theta}}^{+}}$ (size: $K \times p$) and computes the Fisher information matrix ($\mathcal{O}(mK p^2)$). 
    \item Lines 8-9 calculate the Fisher information matrix's inverse and determinant (both $\mathcal{O}(p^3)$).
\end{enumerate}
Operations on lines 6 and 7 within the loop are computational bottlenecks. The sum of $m$ in the loop can reach $L^2$ in the worst case and $L$ in the best case, leading to Alg.~\ref{alg:maye}'s time complexity expressed as:
\begin{equation*}
    T(n) = \mathcal{O}(\sum_{m} m K p^2) = 
    \begin{cases}
        \mathcal{O}(L K p^2) & \textrm{in the best} \\ 
        \mathcal{O}(L^2 K p^2) & \textrm{in the worst} \\
    \end{cases}
\end{equation*}

\section{Proof of Time Complexity of Greedy Algorithm Using Only Initial Calibration Parameters}
\label{appendix:ours}

In this appendix, we derive the time complexity of the greedy algorithm with utility evaluation at initial calibration parameters (Alg.~\ref{alg:ours}). We use the same symbols as introduced in Appendix~\ref{appendix:maye}.

\begin{enumerate}
    \item Line 4 in Alg.~\ref{alg:ours} iterates $L$ times.
    \item Line 5 evaluates Jacobian $\vect{J}(\mathcal{D}^{new})|_{\vect{\theta}^{0}}$ (size: $K \times p$) and computes the Fisher information matrix ($\mathcal{O}(K p^2)$). 
    \item Lines 6-7 calculates the Fisher information matrix's inverse and determinant (both $\mathcal{O}(p^3)$).
    \item After the loop, line 12 conducts nonlinear least-square on $M \times K$ data ($\mathcal{O}(MK \cdot p^2)$).
\end{enumerate}
The operation on line 5 within the loop is a computational bottleneck, leading to Alg.~\ref{alg:ours}'s time complexity expressed as:
\begin{equation*}
    T(n) = \mathcal{O}(L \times K p^2 + (MK) p^2) = \mathcal{O}(L K p^2).
\end{equation*}


\bibliographystyle{IEEEtran}
\bibliography{IEEEabrv,References}

\end{document}


\subsection{How does our calibration solution approach correspond to that of Maye~\cite{maye2016online}?}

In this subsection, it is shown how our calibration solution approach aligns with that of Maye~\cite{maye2016online}. This section is generally based on the careful derivation presented in Maye~\cite{maye2016online}, with a particular focus on how it applies to our calibration formulation and solution approach.
\textcolor{red}{
We also have made necessary adjustments to accommodate our problem formulation, notably relaxing the assumption of identical distribution of data for each time step.
}

The first step of our calibration method is to identify the different variables in the system and establish a probabilistic model through which they interact. These variables shall be classified in the following three categories.
\begin{enumerate}
    \item The data variable denoted by 
    \begin{equation*}
        \vect{X} = 
        \begin{bmatrix} 
            \dotsc {^{I_n}\widetilde{\vect{a}}}^T \, {^{g_n}\widetilde{\vect{\omega}}}^T \dotsc 
        \end{bmatrix}^T,
    \end{equation*}
    where $^{I_n}\widetilde{\vect{a}}$ and ${^{g_n}\widetilde{\vect{\omega}}}$ are accelerometer and gyroscope measurements for each IMU $n \in \{ 0, \dotsc, N \}$. A realization of the observable random variable, $\vect{x}_k$, is associated with the measurements of IMUs at each time $k$.
    \item The calibration variable denoted by 
    \begin{equation*}
        \vect{\Theta} = 
        \begin{bmatrix} 
            {\vect{p}_{{I}_1}}^T \, \dotsc \, {\vect{p}_{{I}_N}}^T \, {\vect{q}_{{I}_1}}^T \, \dotsc \, {\vect{q}_{{I}_N}}^T \, {\vect{q}_{g_0}}^T \, {\vect{q}_{g_1}}^T \, \dotsc \, {\vect{q}_{g_N}}^T 
        \end{bmatrix}^T.
    \end{equation*}
    This latent random variable is the primary object of interest and comprises all the calibration parameters of the IMUs.
    \item The nuisance variable represented by 
    \begin{equation*}
        \vect{\Psi} = 
        \begin{bmatrix} 
            {\dotsc {^{I_0}\vect{\alpha}_{I_0}}^T \, {\vect{b}_{a_{n}}}^T \, {\vect{b}_{g_{n}}}^T \dotsc} 
        \end{bmatrix}^T,
    \end{equation*}
    where $\vect{b}_{a_{n}}, \, \vect{b}_{g_{n}}$ are the time-varying accelerometer and gyroscope biases for each IMU $n \in \{ 0, \dotsc, N \}$ and $^{I_0}\vect{\alpha}_{I_0} := ^{I_0}\vect{\alpha}$
    the angular acceleration of the base IMU. A realization of the nuisance random variable, $\vect{\psi}_k$, is associated with each time $k$.
    This latent random variable is not of direct interest, but the data distribution depends on it.
\end{enumerate}

Assuming a set of realizations for the observable random variables $\vect{x}_1, \dotsc, \vect{x}_K$, \textcolor{red}{which are independent but not necessarily identically distributed}, and another set for the nuisance random variables $\vect{\psi}_1, \dotsc, \vect{\psi}_K$, also independent and not necessarily identically distributed. We will denote these sets as $\vect{x}_{1:K}$ and $\vect{\psi}_{1:K}$ (or, just simply, $\vect{\psi}$), along with the realization of calibration variable $\vect{\theta}$. The posterior joint probability density function of the latent variables are then given by
\begin{equation} \label{eq:joint pdf of psi and phi}
\begin{split}
    p(\vect{\psi}, \vect{\theta} | \vect{x}_{1:K}) 
    &= \frac{p(\vect{x}_{1:K} | \vect{\psi}_{1:K}, \vect{\theta}) \cdot p(\vect{\psi}_{1:K}, \vect{\theta})}{p(\vect{x}_{1:K})} \\
    &= \eta \cdot p(\vect{\psi}_1) \prod_{k=2}^{K} p(\vect{\psi}_k|\vect{\psi}_{k-1}, \vect{\theta}) \cdot \prod_{k=1}^{K} p(\vect{x}_k|\vect{\psi}_{k}, \vect{\theta}).
\end{split}
\end{equation}
This is Bayes' rule, and those interested in further details are encouraged to refer to Chapter 11.4 of Thrun et al.~\cite{thrun2002probabilistic}.
As our final goal is to perform inference on the calibration variable, $\vect{\theta}$, we shall marginalize over the nuisance variables, $\vect{\psi}$, to obtain the marginal posterior density function
\begin{equation} \label{eq:marginal posterior pdf}
    p(\vect{\theta}|\vect{x}_{1:K}) = \int p(\vect{\psi}, \vect{\theta} | \vect{x}_{1:K}) d \vect{\psi}.
\end{equation}
\textcolor{red}{In the field of probabilistic robotics, it is a common practice to make a Gaussian noise assumption in both measurement and motion models, as noted by Thrun et al.~\cite{thrun2002probabilistic}.
Therefore, given this assumption, the posterior density in Eq.~\ref{eq:joint pdf of psi and phi} follows a normal distribution given the product of probability density functions of normal distributions}
\begin{equation} \label{eq:marginal joint posterior pdf}
    p(\vect{\psi}, \vect{\theta} | \vect{x}_{1:K}) = \mathcal{N}(\vect{\mu}_{\vect{\psi} \vect{\theta}}, {\vect{\mathcal{I}}(\vect{\psi} \vect{\theta}| \vect{x}_{1:K})}^{-1}),
\end{equation}
where $\vect{\mu}_{\vect{\psi} \vect{\theta}}$ is the mean of the joint posterior $\vect{\psi} \vect{\theta}$  and ${\vect{\mathcal{I}}(\vect{\psi} \vect{\theta}| \vect{x}_{1:K})}$ the Fisher information matrix for the joint posterior $\vect{\psi} \vect{\theta}$ --- the negative of the Hessian matrix of the log-likelihood function --- associated with the data variables $\vect{x}_{1:K}$. These values are determined through the solution of a least-squares problem and the computation of the corresponding Hessian, as described in Eq. 3 of the manuscript.

Using the properties of the normal distribution~\cite{bishop2006pattern}, we can derive a closed-form expression for the posterior marginal density in Eq.~\ref{eq:marginal posterior pdf} as
\begin{equation}
    p(\vect{\theta}|\vect{x}_{1:K}) \approx \mathcal{N} (\vect{\mu}_{\vect{\Theta}|\vect{x}_{1:K}}, \vect{\Sigma}_{\vect{\Theta}|\vect{x}_{1:K}})
\end{equation}
where $\vect{\mu}_{\vect{\Theta}|\vect{x}_{1:K}}$ and $\vect{\Sigma}_{\vect{\Theta}|\vect{x}_{1:K}}$ represent partitions of the joint mean and covariance matrix in Eq.~\ref{eq:marginal joint posterior pdf}, respectively.

\subsection{What do covariances for the calibration parameters $\vect{\Theta}$ associated with $\mathcal{D}^{info}$ and $\mathcal{D}^{info},\mathcal{D}^{new}$ mean?}

$\vect{\Sigma}_{\vect{\Theta}|\mathcal{D}^{info}}$ and $\vect{\Sigma}_{\vect{\Theta}|\mathcal{D}^{info},\mathcal{D}^{new}}$ (what are referred to as $\vect{\Sigma}^{prior}$ and $\vect{\Sigma}^{post}$ in Maye's journal paper~\cite{maye2016online}) are the covariance matrices for the calibration parameters $\vect{\Theta}$ associated with a set of data variables $\mathcal{D}^{info}$ and both sets of data variables $\mathcal{D}^{info}$ and $\mathcal{D}^{new}$, respectively. 
These matrices are obtained by calculating the Fisher information matrices, denoted as $\vect{\mathcal{I}}(\vect{\psi} \vect{\theta}| \mathcal{D}^{info})$ and $\vect{\mathcal{I}}(\vect{\psi} \vect{\theta}| \mathcal{D}^{info},\mathcal{D}^{new})$, using the Jacobians of the stacked error terms associated with the dataset
$\mathcal{D}^{info}$ and the combined dataset of $\mathcal{D}^{info}$ and $\mathcal{D}^{new}$ (as outlined in Eq. 13. of Maye~\cite{maye2016online}.) Subsequently, the inverses of these Fisher information matrices are computed, and specific block entries corresponding to the calibration parameters $\vect{\Theta}$ are extracted (as outlined in Eq. 3 and Eq. 4 of Maye~\cite{maye2016online}.) 

\subsubsection{$\vect{\Sigma}_{\vect{\Theta}|\mathcal{D}^{info}}$}

The Fisher information matrix associated with a set of data variables $\mathcal{D}^{info}$ is:
\begin{align} \label{eq:Fisher information 1:K}
\begin{split}
    \vect{\mathcal{I}}(\vect{\psi} \vect{\theta}| \mathcal{D}^{info}) 
    &= {\vect{J}(\mathcal{D}^{info})}^T {\vect{J}(\mathcal{D}^{info})} \\
    &= 
    \begin{bmatrix}
        {\vect{J}_{\vect{\psi}}(\mathcal{D}^{info})}^T \\
        {\vect{J}_{\vect{\theta}}(\mathcal{D}^{info})}^T
    \end{bmatrix}
    \begin{bmatrix}
        \vect{J}_{\vect{\psi}}(\mathcal{D}^{info}) & \vect{J}_{\vect{\theta}}(\mathcal{D}^{info})
    \end{bmatrix} \\
    &= 
    \begin{bmatrix}
        {\vect{J}_{\vect{\psi}}(\mathcal{D}^{info})}^T \vect{J}_{\vect{\psi}}(\mathcal{D}^{info}) & {\vect{J}_{\vect{\psi}}(\mathcal{D}^{info})}^T \vect{J}_{\vect{\theta}}(\mathcal{D}^{info}) \\
        {\vect{J}_{\vect{\theta}}(\mathcal{D}^{info})}^T \vect{J}_{\vect{\psi}}(\mathcal{D}^{info}) & {\vect{J}_{\vect{\theta}}(\mathcal{D}^{info})}^T \vect{J}_{\vect{\theta}}(\mathcal{D}^{info})
    \end{bmatrix},
\end{split}
\end{align}
where $\vect{J}(\mathcal{D}^{info})$ is the Jacobian matrix of the stacked error terms associated with $\mathcal{D}^{info}$.
After taking the inverse of the Fisher information matrix $\vect{\mathcal{I}}(\vect{\psi} \vect{\theta}| \mathcal{D}^{info})$ in Eq.~(\ref{eq:Fisher information 1:K}), we pick out the block entries corresponding to $\vect{\Theta}$, which is at the lower-right block, to obtain the covariance matrix of $\vect{\Theta}$:
\begin{equation} \label{eq: covariance 1:K}
    \vect{\Sigma}_{\vect{\Theta}|\mathcal{D}^{info}} = \big[ {\vect{\mathcal{I}}(\vect{\psi} \vect{\theta}| \mathcal{D}^{info})}^{-1} \big]_{\vect{\Theta}\vect{\Theta}}. 
\end{equation}
For your information, the written-out form of $\vect{\Sigma}_{\vect{\Theta}|\mathcal{D}^{info}}$ is:
\begin{equation}
    ({\vect{J}_{\vect{\theta}}}^{T} {\vect{J}_{\vect{\theta}}})^{-1} + ({\vect{J}_{\vect{\theta}}}^{T} {\vect{J}_{\vect{\theta}}})^{-1} {\vect{J}_{\vect{\theta}}}^{T} {\vect{J}_{\vect{\psi}}} ({\vect{J}_{\vect{\psi}}}^{T} {\vect{J}_{\vect{\psi}}} - {\vect{J}_{\vect{\psi}}}^{T} {\vect{J}_{\vect{\theta}}} ({\vect{J}_{\vect{\theta}}}^{T} {\vect{J}_{\vect{\theta}}})^{-1} {\vect{J}_{\vect{\theta}}}^{T} {\vect{J}_{\vect{\psi}}})^{-1} {\vect{J}_{\vect{\psi}}}^{T} {\vect{J}_{\vect{\theta}}} ({\vect{J}_{\vect{\theta}}}^{T} {\vect{J}_{\vect{\theta}}})^{-1}.
\end{equation}
For convenience, $\mathcal{D}^{info}$ is omitted. Be aware that this is different from $({\vect{J}_{\vect{\theta}}}^T \vect{J}_{\vect{\theta}})^{-1}$, which is easily thought to be $\vect{\Sigma}_{\vect{\Theta}|\mathcal{D}^{info}}$.
Though the written-out form is complicated, the bottom line is that the inversion of the Fisher information matrix $\vect{\mathcal{I}}(\vect{\psi} \vect{\theta}| \mathcal{D}^{info})$ is performed, and its lower-right block corresponding to $\vect{\Theta}$ is chosen.


\subsubsection{$\vect{\Sigma}_{\vect{\Theta}|\mathcal{D}^{info},\mathcal{D}^{new}}$}

The Fisher information matrix associated with both sets of data variables $\mathcal{D}^{info}$, $\mathcal{D}^{new}$ is:
\begin{align} \label{eq:Fisher information 1:2K}
\begin{split}
    \vect{\mathcal{I}}(\vect{\psi} \vect{\theta}| \mathcal{D}^{info},\mathcal{D}^{new}) 
    &= {\vect{J}(\mathcal{D}^{info},\mathcal{D}^{new})}^T {\vect{J}(\mathcal{D}^{info},\mathcal{D}^{new})} \\
    &= 
    \begin{bmatrix}
    {\vect{J}(\mathcal{D}^{info})}^T & {\vect{J}(\mathcal{D}^{new})}^T
    \end{bmatrix}
    \begin{bmatrix}
    {\vect{J}(\mathcal{D}^{info})} \\ {\vect{J}(\mathcal{D}^{new})}
    \end{bmatrix} \\
    &= 
    \begin{bmatrix}
        {\vect{J}_{\vect{\psi}}(\mathcal{D}^{info})}^T & {\vect{J}_{\vect{\psi}}(\mathcal{D}^{new})}^T \\
        {\vect{J}_{\vect{\theta}}(\mathcal{D}^{info})}^T & {\vect{J}_{\vect{\theta}}(\mathcal{D}^{new})}^T
    \end{bmatrix}
    \begin{bmatrix}
        \vect{J}_{\vect{\psi}}(\mathcal{D}^{info}) & \vect{J}_{\vect{\theta}}(\mathcal{D}^{info}) \\
        \vect{J}_{\vect{\psi}}(\mathcal{D}^{new}) & \vect{J}_{\vect{\theta}}(\mathcal{D}^{new})
    \end{bmatrix} \\
    &= 
    \begin{bmatrix}
        {\vect{J}_{\vect{\psi}}(\mathcal{D}^{info})}^T \vect{J}_{\vect{\psi}}(\mathcal{D}^{info}) + {\vect{J}_{\vect{\psi}}(\mathcal{D}^{new})}^T \vect{J}_{\vect{\psi}}(\mathcal{D}^{new}) & {\vect{J}_{\vect{\psi}}(\mathcal{D}^{info})}^T \vect{J}_{\vect{\theta}}(\mathcal{D}^{info}) + {\vect{J}_{\vect{\psi}}(\mathcal{D}^{new})}^T \vect{J}_{\vect{\theta}}(\mathcal{D}^{new}) \\
        {\vect{J}_{\vect{\theta}}(\mathcal{D}^{info})}^T \vect{J}_{\vect{\psi}}(\mathcal{D}^{info}) + {\vect{J}_{\vect{\theta}}(\mathcal{D}^{new})}^T \vect{J}_{\vect{\psi}}(\mathcal{D}^{new}) & {\vect{J}_{\vect{\theta}}(\mathcal{D}^{info})}^T \vect{J}_{\vect{\theta}}(\mathcal{D}^{info}) + {\vect{J}_{\vect{\theta}}(\mathcal{D}^{new})}^T \vect{J}_{\vect{\theta}}(\mathcal{D}^{new})
    \end{bmatrix} \\
    & = \vect{\mathcal{I}}(\vect{\psi} \vect{\theta}| \mathcal{D}^{info}) + \vect{\mathcal{I}}(\vect{\psi} \vect{\theta}| \mathcal{D}^{new}),
\end{split}
\end{align}
where $\vect{J}(\mathcal{D}^{info},\mathcal{D}^{new})$ is the Jacobian matrix of the stacked error terms associated with $\mathcal{D}^{info},\mathcal{D}^{new}$.

After taking the inverse of the Fisher information matrix $\vect{\mathcal{I}}(\vect{\psi} \vect{\theta}| \mathcal{D}^{info},\mathcal{D}^{new})$ in Eq.~(\ref{eq:Fisher information 1:2K}), now we pick out the block entries corresponding to $\vect{\Theta}$, which is in the lower-right block, to obtain the covariance matrix of $\vect{\Theta}$:
\begin{equation} \label{eq: covariance 1:2K}
\begin{split}
    \vect{\Sigma}_{\vect{\Theta}|\mathcal{D}^{info},\mathcal{D}^{new}} = \big[ {\vect{\mathcal{I}}(\vect{\psi} \vect{\theta}| \mathcal{D}^{info},\mathcal{D}^{new})}^{-1} \big]_{\vect{\Theta}\vect{\Theta}}.
\end{split}
\end{equation}

\subsection{How are our method and Maye's method different?}

Eq.~\ref{eq: covariance 1:K} and Eq.~\ref{eq: covariance 1:2K} suggest that the computation of the covariance for $\vect{\Theta}$ depends on the specific value of $\vect{\theta}$ at which the Jacobians, and consequently, the Fisher information matrices in Eq.\ref{eq:Fisher information 1:K} and Eq.\ref{eq:Fisher information 1:2K} are calculated.

Maye~\cite{maye2013self,maye2016online} employs the value of $\vect{\theta}$ as the ``current'' estimate for each iteration (as indicated by the phrase `where J(t) is the Jacobian matrix of the stacked error terms evaluated at the current estimate,' located just below Eq. 9 in the journal~\cite{maye2016online} that clarifies which value is used for evaluating Jacobians, Fisher information, and covariance). By introducing the notation of the current estimate for each iteration involving the existing informative dataset $\mathcal{D}^{info}$ and the new dataset under examination $\mathcal{D}^{new}$: $\widehat{\vect{\theta}}|{\mathcal{D}^{info},\mathcal{D}^{new}}$, the computed Fisher information matrix is:
\begin{align} \label{eq:Maye}
\begin{split}
    \vect{\mathcal{I}}(\vect{\psi} \vect{\theta}| \mathcal{D}^{info}, \mathcal{D}^{new}) |_{\vect{\theta} = \widehat{\vect{\theta}}|{\mathcal{D}^{info},\mathcal{D}^{new}}} 
    &= \vect{\mathcal{I}}(\vect{\psi} \vect{\theta}| \mathcal{D}^{info}) |_{\vect{\theta} = \widehat{\vect{\theta}}|{\mathcal{D}^{info},\mathcal{D}^{new}}} + \vect{\mathcal{I}}(\vect{\psi} \vect{\theta}| \mathcal{D}^{new}) |_{\vect{\theta} = \widehat{\vect{\theta}}|{\mathcal{D}^{info},\mathcal{D}^{new}}} \\
    &= {\vect{J}(\mathcal{D}^{info})}|^{T}_{\vect{\theta} = \widehat{\vect{\theta}}|{\mathcal{D}^{info},\mathcal{D}^{new}}} {\vect{J}(\mathcal{D}^{info})}|_{\vect{\theta} = \widehat{\vect{\theta}}|{\mathcal{D}^{info},\mathcal{D}^{new}}} + {\vect{J}(\mathcal{D}^{new})}|^{T}_{\vect{\theta} = \widehat{\vect{\theta}}|{\mathcal{D}^{info},\mathcal{D}^{new}}} {\vect{J}(\mathcal{D}^{new})}|_{\vect{\theta} = \widehat{\vect{\theta}}|{\mathcal{D}^{info},\mathcal{D}^{new}}} \\
    .
\end{split}
\end{align}

Our approach employs the value of $\vect{\theta}$ as the uncalibrated value $\vect{\theta}^0$ for all iterations. The computed Fisher information matrix is:
\begin{align} \label{eq:ours}
\begin{split}
    \vect{\mathcal{I}}(\vect{\psi} \vect{\theta}| \mathcal{D}^{info}, \mathcal{D}^{new}) |_{\vect{\theta} = \vect{\theta}^0}
    &= \vect{\mathcal{I}}(\vect{\psi} \vect{\theta}| \mathcal{D}^{info}) |_{\vect{\theta} = \vect{\theta}^0} + \vect{\mathcal{I}}(\vect{\psi} \vect{\theta}| \mathcal{D}^{new}) |_{\vect{\theta} = \vect{\theta}^0} \\
    &= {\vect{J}(\mathcal{D}^{info})}|^{T}_{\vect{\theta} = \vect{\theta}^0} {\vect{J}(\mathcal{D}^{info})}|_{\vect{\theta} = \vect{\theta}^0} + {\vect{J}(\mathcal{D}^{new})}|^{T}_{\vect{\theta} = \vect{\theta}^0} {\vect{J}(\mathcal{D}^{new})}|_{\vect{\theta} = \vect{\theta}^0} \\
    .
\end{split}
\end{align}

In either case, it remains true that the estimated Fisher information matrix needs to be inverted, and the block entries corresponding to $\vect{\Theta}$ are selected. However, our modification in Eq.~\ref{eq:ours} eliminates the need to update $\vect{\mathcal{I}}(\vect{\psi} \vect{\theta}| \mathcal{D}^{info}) |_{\vect{\theta} = \vect{\theta}^0}$ every time a new $\mathcal{D}^{new}$ is introduced, whereas Maye's approach~\ref{eq:Maye} requires the computation of a new $\vect{\mathcal{I}}(\vect{\psi} \vect{\theta}| \mathcal{D}^{info}) |_{\vect{\theta} = \widehat{\vect{\theta}}|{\mathcal{D}^{info},\mathcal{D}^{new}}}$ whenever a new $\mathcal{D}^{new}$ is introduced.

\subsection{Where does Maye et al.~\cite{maye2016online} deviate from its assumption of independent and identically distributed data variables at each time step?}

Maye et al.~\cite{maye2016online} begins its discussion in Section 3.1 by asserting that the data variables at each time step are presumed to be \emph{independent and identically distributed}, as illustrated below:
\begin{quote}
\begin{mdframed}
Assuming a realization $\vect{x}_1, \dotsc, \vect{x}_N$ of an i.i.d. data sample, henceforth denoted by $\vect{x}_{1:N}$,
\dots
\end{mdframed}
\end{quote}
However, this assumption stands in contrast to the subsequent experimental setup presented in the paper. In the following paragraphs, it becomes evident that the experimental verifications outlined in the paper itself do not adhere to this assumption, thereby challenging the validity of the statement mentioned earlier.

Refer to Section 4.1 in Maye et al.~\cite{maye2016online}, which presents the 2-D SLAM problem involving a wheel odometry sensor and a laser rangefinder (LRF). In this scenario, the relative pose between the LRF frame and the robot frame (which is colocated with the wheel odometry sensor) remains unknown and is going to be estimated:
\begin{quote}
\begin{mdframed}
We consider a differential-drive wheeled robot equipped with a LRF. \emph{While moving in a plane}, the LRF measures distances to poles disseminated on the ground. Furthermore, the robot is endowed with odometry sensors, measuring its linear and angular velocity. This scenario is a prototypical SLAM problem. Freely adopting the SLAM jargon, we refer to the poles as landmarks.
\end{mdframed}
\end{quote}
In their subsection ``Estimation Model,'' they state:
\begin{quote}
\begin{mdframed}
With regard to Section 3, \emph{a data sample of size $N$ consists in $\vect{v}^{o}_{1:N}$, $^{l}\vect{l}_{1:Nl_{1:N}}$}, the calibration variable in $\vect{\Theta} = \{ ^{r}\vect{T}_{l} \}$, and the nuisance variable in $\vect{\Psi} = \{ ^{w}\vect{T}_{r_{1:N}}, ^{w}\vect{l}_{1:Nl} \} $.
\end{mdframed}
\end{quote}
where $\vect{v}$ and $^{l}\vect{l}$ represent wheel odometry and LRF measurements, respectively. Additionally, $N$ notates the number of time steps, and $Nl$ denotes the number of landmarks detected by the LRF at each time step. As previously mentioned, these measurements are obtained while the robot is in motion, implying that $\vect{v}_k$ and $^{l}\vect{l}_{1:Nl_k}$ for each time step $k$ do not conform to the assumption of identical distribution.

Now, see section 4.2. in Maye et al.~\cite{maye2016online}, which articulates the extrinsic calibration of a visual-inertial sensor unit.  
\begin{quote}
\begin{mdframed}
In this experiment we considered the extrinsic calibration of a visual-inertial sensor unit, which consisted of two MT9V034 global shutter cameras in a plan-parallel config- uration and an ADIS16448 IMU (see Figure 12 and an extended description of the device in Nikolic et al. (2014)).

\dots

An AprilTag calibration pattern board was placed in front of the setup in order to provide a set of fixed landmarks for the cameras to observe \emph{while in motion} (Furgale et al., 2013a; Olson, 2011). \emph{With the sensor unit fixed to the mount, we collected data while performing two different motion patterns.} Then, we released the unit from the mount and performed the third motion pattern.
\end{mdframed}
\end{quote}
Although this section does not explicitly specify the data variables acquired from either the cameras or the Inertial Measurement Unit (IMU), it implies that the data was collected while the sensor unit was in motion, too. This suggests that neither the data from the cameras nor the IMU conforms to the assumption of identical distribution at each time step.

\textcolor{red}{
[NOTE] This section does not intend to point out any flaws in Maye~\cite{maye2016online}'s meticulous proofs throughout Section 3.1. However, it highlights a discrepancy between the underlying assumption of its experimental setups (in Section 4.1. and 4.2.) and the assertions made in its proofs. While the proofs assume that data for each time step are independent and identically distributed, the experimental data do not adhere to this assumption, as each time step's data does not conform to identical distribution.
}




\bibliographystyle{IEEEtran}
\bibliography{IEEEabrv,References}